\documentclass[10pt,twocolumn,letterpaper]{article}

\usepackage[pagenumbers]{cvpr} 

\renewcommand{\paragraph}[1]{\vspace{0.5em}\noindent\textbf{#1}}
\newcommand{\paragraphtight}[1]{\noindent\textbf{#1}}

\usepackage{enumitem}
\usepackage{mathrsfs}   
\usepackage{amsmath,amssymb}
\usepackage{graphicx, wrapfig, float, indentfirst}
\usepackage{subcaption}

\usepackage[linesnumbered,ruled,vlined]{algorithm2e} 
\makeatletter
\newcommand{\removelatexerror}{\let\@latex@error\@gobble}
\makeatother
\usepackage{mathrsfs}   

\usepackage[table]{xcolor}
\usepackage{tabularx}
\usepackage{multirow}
\usepackage{array}
\usepackage{booktabs,tabularx,array,multirow}
\newcolumntype{Y}{>{\centering\arraybackslash}X}
\usepackage{subcaption}  
\usepackage{pifont} 
\newcommand{\cmark}{\ding{51}} 
\newcommand{\xmark}{\ding{55}} 

\usepackage[accsupp]{axessibility}  

\definecolor{cvprblue}{rgb}{0.21,0.49,0.74}
\usepackage[pagebackref,breaklinks,colorlinks,allcolors=cvprblue]{hyperref}

\def\paperID{*****} 
\def\confName{CVPR}
\def\confYear{2026}

\title{AdaRadar: Rate Adaptive Spectral Compression for Radar-based Perception}

\usepackage{orcidlink}
\author{
Jinho Park$^{1}$\orcidlink{0000-0003-1613-0769} \qquad
Se Young Chun$^{2}$\orcidlink{0000-0001-8739-8960} \qquad
Mingoo Seok$^{1}$\orcidlink{0000-0002-9722-0979} \\
$^{1}$Columbia University \qquad $^{2}$Seoul National University \\
{\tt\small jp4327@columbia.edu, sychun@snu.ac.kr, ms4415@columbia.edu} \\
\href{https://jp4327.github.io/adaradar/}{\tt\small{jp4327.github.io/adaradar/}}
}

\begin{document}
\maketitle
\begin{abstract}
Radar is a critical perception modality in autonomous driving systems due to its all-weather characteristics and ability to measure range and Doppler velocity.
However, the sheer volume of high-dimensional raw radar data saturates the communication link to the computing engine (e.g., an NPU), which is often a low-bandwidth interface with a data rate provisioned for only a few low-resolution range–Doppler frames.
A generalized codec for utilizing high-dimensional radar data is notably absent, while existing image-domain approaches are unsuitable, as they typically operate at fixed compression ratios and fail to adapt to varying or adversarial conditions.
In light of this, we propose radar data compression with adaptive feedback.
It dynamically adjusts the compression ratio by performing gradient descent from the proxy gradient of detection confidence with respect to the compression rate.
We employ a zeroth-order gradient approximation, which enables gradient computation even with non-differentiable core operations such as pruning and quantization.
This also avoids transmitting the gradient tensors over the band-limited link, which, if estimated, would be as large as the original radar data.
In addition, we have found that radar feature maps are heavily concentrated on a few frequency components. Thus, we apply the discrete cosine transform to the radar data cubes and selectively prune the coefficients. 
We preserve the dynamic range of each radar patch through scaled quantization.
Combining those techniques, our proposed online adaptive compression scheme achieves over 100\texttimes{} feature size reduction at minimal performance drop (\textasciitilde{}1\%p). 
We validate our results on the RADIal, CARRADA, and Radatron datasets.
\end{abstract}    
\section{Introduction} \label{sec:intro}

\begin{figure}[t] 
	\centering
	\includegraphics[width=0.95\linewidth]{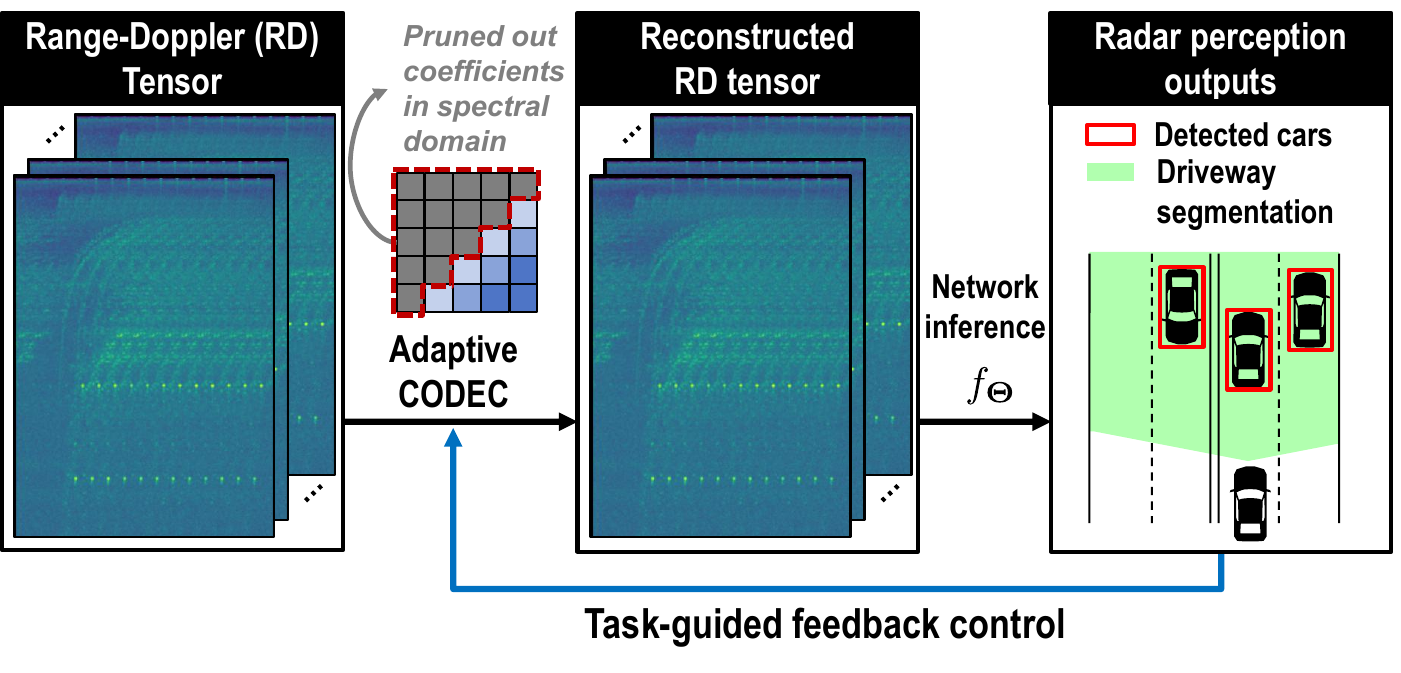}
        \vspace{-1em}
        \caption{\textbf{Adaptive codec with task-guided feedback control.} 
        We propose an adaptive codec that compresses high-dimensional range-Doppler data by pruning spectral domain coefficients. A feedback loop adaptively regulates the compression ratio, guided by the performance of downstream perception tasks.
        }
        \label{fig:abs}
        \vspace{-1em}  
\end{figure}

\noindent Autonomous driving systems incorporate radar as a crucial component of their perception stack.
Compared to camera and LiDAR counterparts, it delivers dependable performance across all weather and lighting conditions, along with accurate physical measurements such as range and Doppler velocity of surrounding objects \cite{bijelic2020seeing, paek2022k, rebut2022raw, yang2020radarnet}.
Such innovation has been driven by the emergence of low-cost millimeter-wave (mmWave) radar system-on-chip (SoC) platforms \cite{TI_IWR1843_2024}.
Previous studies leveraged the rich information contained in raw radar tensors, enabled by modern multiple-input multiple-output (MIMO) techniques, by directly using them as network inputs rather than relying on point clouds \cite{rebut2022raw, paek2022k}.

Modern MIMO radar sensors achieve high-resolution imaging through dense sampling and a large number of transceiver pairs.
However, unlike camera images with only three RGB channels, the data volume in MIMO radars scales quadratically with the number of transmit (Tx) and receive (Rx) antennas. 
For instance, a 16\texttimes{}12 cascaded MIMO radar system \cite{TI_IWR1843_2024} configured with 512 range bins and 256 Doppler bins produces roughly 100 MB per frame, corresponding to a multi-Gbps throughput at 10 fps.
The immense volume of radar tensors creates a significant bottleneck in the sensor-to-compute link, which connects the radar front-end with the downstream processor.
This link is often low-bandwidth, such as the Controller Area Network (CAN) bus \cite{TI_TCAN3414_2023}, with a limited throughput of a few Mbps.

\begin{figure*}[t] 
	\centering
	\includegraphics[width=0.95\linewidth]{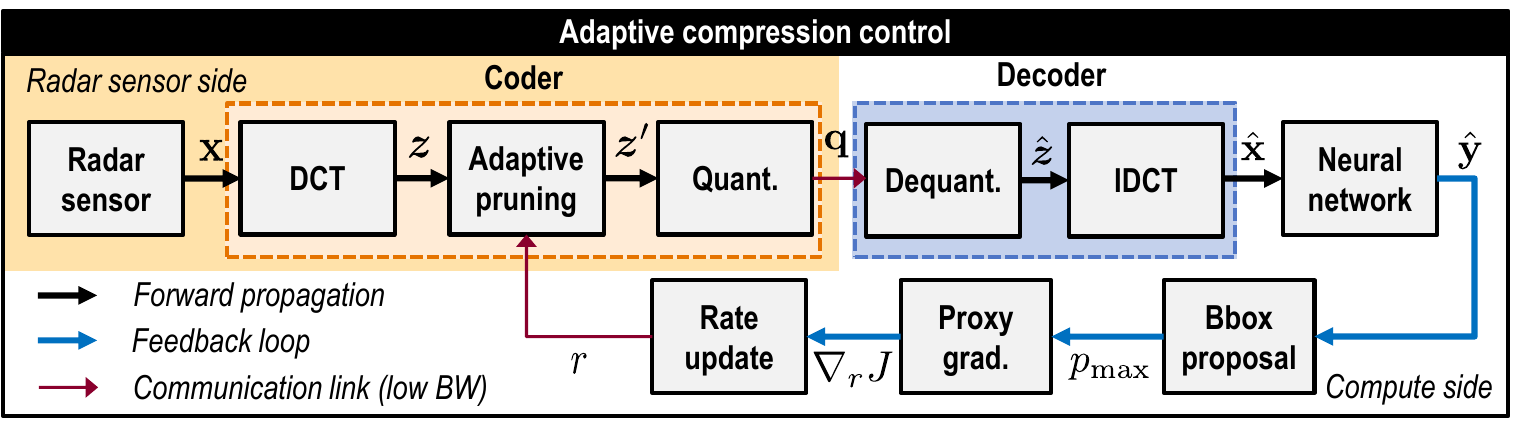}
    \caption{\textbf{AdaRadar: Online rate-adaptive radar compression framework.} Our proposed method introduces a feedback loop in which the proxy gradient is computed from the detection outputs to update the compression ratio adaptively. This avoids the need for backpropagation through the communication channel. The radar tensor is compressed using DCT, adaptive spectral pruning, and scaled quantization, then transmitted to the compute side. In an object-detection setting, the neural network produces detection results from decompressed radar data cubes. The loop uses proposed bounding boxes to estimate the proxy gradient, thereby updating the pruning rate.}
	\label{fig:proposed}
    \vspace{-1em}  
\end{figure*}

Despite the multi-dimensional nature of radar features, which creates a severe communication bottleneck, only limited efforts have been made to design a dedicated coder–decoder (codec) for compressed radar feature representations.
In computer vision, codecs such as JPEG \cite{pennebaker1992jpeg} effectively reduce image size for storage and transmission while preserving perceptual quality.
In contrast, some recent radar compression methods often rely on the constant false alarm rate (CFAR) algorithm to compress raw signals into sparse point clouds \cite{caesar2020nuscenes}, which inevitably leads to information loss and degraded downstream performance \cite{paek2022k}.
RadarOcc \cite{ding2024radarocc} proposed an energy-based compression scheme with deformable attention to mitigate information loss from retaining the top-$K$ index–value pairs across Doppler, azimuth, and elevation dimensions.

Furthermore, none of these approaches provides an \textit{adaptive} codec capable of dynamically adjusting the compression ratio at test time--an essential feature for maintaining performance under varying or adversarial conditions. Existing codecs with fixed compression rates are inherently inefficient, either wasting bandwidth when underutilized or suffering performance degradation when over-compressed.

In light of this, we propose \textbf{AdaRadar}, an \textit{adaptive} compression framework that dynamically modulates the compression rate based on task feedback as illustrated in Fig.~\ref{fig:proposed}.
It leverages this feedback loop to compute a task-aware gradient without requiring full backpropagation.
This is critical for two reasons: 1) our core operations--pruning and quantization--are non-differentiable, and 2) transmitting a standard gradient approximation would be as large as the original radar data, nullifying all compression gains. 
AdaRadar bypasses this bottleneck by using a zeroth-order gradient estimator to compute a lightweight proxy gradient. 
Our framework generalizes across different radar perception tasks, including object detection and semantic segmentation, by requiring only post-hoc confidence outputs from the downstream model.

We find that radar signals exhibit strong sparsity in the spectral domain, a property we exploit to achieve substantial compression without loss of task performance.
Its rank-based pruning mechanism adaptively compresses features with varying frequency structures, preserving the most informative components.
Our codec generalizes across \textit{any} frequency-modulated continuous-wave (FMCW) radar by employing scaled quantization with a per-patch scaling factor, thereby preserving a high dynamic range at fine granularity.
The rank-based adaptive pruning allows compression among different frequency structures.
AdaRadar can be seamlessly integrated into existing radar systems, as the sensor-side requires only lightweight algorithmic kernels, e.g., the discrete cosine transform (DCT), sorting, and rounding, which are readily supported by embedded DSPs.

In summary, the contribution of our work is threefold:
\begin{itemize}
    \setlength\parsep{0pt}
    
    \item We present a proxy-gradient-based adaptive control method that utilizes only forward passes. Our method achieves over 100\texttimes reduction in radar data volume while causing a negligible drop (\textasciitilde{}1\%p) in downstream task performance.
    
    \item We show that radar data is highly compressible in the spectral domain since its information is heavily clustered within a few frequencies (e.g., high frequencies for the RADIal dataset \cite{rebut2022raw}). Our generalized codec leverages this structure, significantly reducing the data load on the communication link. This method can be annexed to existing DNN frameworks without any network modification.

    \item We evaluate our method on multiple radar perception tasks across RADIal \cite{rebut2022raw}, CARRADA \cite{ouaknine2021carrada}, and Radatron \cite{radatron} datasets, featuring object detection and semantic segmentation. The results show robust generalization to both urban and freeway environments.
    
\end{itemize}
    
\section{Related works} \label{sec:related}

\subsection{Radar-based perception systems}

\paragraphtight{Radar as sensing modality.}
Radar has become a vital sensing modality in autonomous driving due to its all-weather and all-light detection capabilities compared to camera and LiDAR alternatives.
Camera is a baseline modality in autonomous vehicles; however, it is prone to illumination change and can be easily occluded by rain and snow particles \cite{hendrycks2019benchmarking, park2024ul}.
While LiDAR provides high-resolution, centimeter-accurate 3D point clouds, it remains vulnerable in poor weather because its operating wavelengths between 750nm–1.5\textmu{}m are too short to penetrate precipitation \cite{kurup2021dsor}.
Radar is a cost-effective sensing modality that remains unaffected by the aforementioned lighting and weather conditions, as it actively emits radio waves of \textasciitilde{}4mm in wavelength and measures their reflections \cite{abdu2021application}.

\begin{figure}[t] 
	\centering
	\includegraphics[width=1\linewidth]{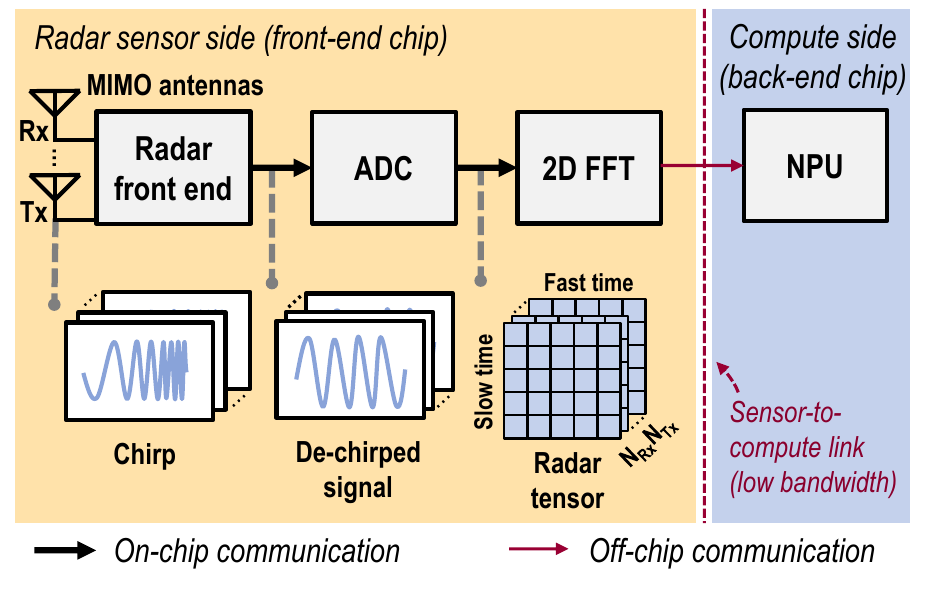}
        \vspace{-2em}
        \caption{\textbf{Radar-based perception system overview.} 
        An FMCW radar transmits linearly swept‑frequency chirps. The incoming echo is mixed with a copy of the transmitted chirp at the receiver, yielding a de‑chirped intermediate‑frequency signal that the ADC digitizes to form a raw radar tensor. Successive FFTs along the fast‑time and slow‑time axes convert this tensor into a range–Doppler cube. The large, raw radar tensor is transferred to the NPU over the power-hungry sensor-to-compute link for network inference.}
        \label{fig:overview}
        \vspace{-1em}  
\end{figure}

\paragraph{Radar systems.} 
FMCW radars \cite{TI_IWR1843_2024} produce range-Doppler radar data cubes that play an analogous role to pixel grids in images.
A typical radar front-end chip, such as \cite{TI_IWR1843_2024}, consists of a radar RF circuit, an analog-to-digital converter (ADC), and a digital signal processor (DSP) computing the fast Fourier transform (FFT) (Fig.~\ref{fig:overview}).
Chirps reflected by surrounding objects are captured by the Rx antenna and mixed with a copy of the transmitted chirp, yielding a de‑chirped intermediate‑frequency (IF) signal \cite{richards2005fundamentals}.
Such a process is carried out multiple times with combinations of Tx and Rx pairs.
The IF waveform is digitized by the ADC and subsequently processed by a two‑dimensional FFT.
The result is a three‑dimensional data cube with dimensions of fast time, slow time, and virtual antenna pairs, respectively encoding range, Doppler velocity, and azimuth‑elevation information.
The back-end chip is typically a SoC that integrates a central processing unit (CPU) and a neural processing unit (NPU) to run ML models, such as object tracking. 

\paragraph{Energy bottleneck.}
The energy bottleneck of modern processors, including recent AI accelerators, is \textit{communication}, not computation \cite{jouppi2017datacenter, chen2016eyeriss, park2022bin}.
For example, off-chip memory access consumes orders of magnitude more energy and latency than \textit{on}-chip memory access, far exceeding the cost of computation itself \cite{horowitz20141, lee2021link}.
This is pertinent to autonomous driving systems with radar sensors \cite{TI_IWR1843_2024} that utilize data links, such as the CAN bus \cite{TI_TCAN3414_2023}, which often exhibit a similar or higher energy cost per bit for data access compared to off-chip DRAM due to a longer wire distance.

\paragraph{Raw radar tensor.} 
The use of radar as an essential detection modality for autonomous driving is highlighted in \cite{major2019vehicle}.
Inspired by deep neural networks that leverage raw feature inputs rather than handcrafted features \cite{krizhevsky2012imagenet}, researchers have evaluated raw radar tensors with vision-domain architectures \cite{rebut2022raw}.
In line with this, numerous raw radar tensor-based datasets have been released \cite{ouaknine2021carrada, zhang2021raddet, mostajabi2020high, sheeny2021radiate, wang2021rethinking, radatron}.
However, a single radar tensor frame in \cite{paek2022k} occupies approximately 500 MB, leading to raw data sizes exceeding several terabytes in widely used datasets \cite{paek2022k, ouaknine2021carrada}.
Communication of these raw tensors poses a non-negligible link budget.

\subsection{Feature compression}

\paragraphtight{Radar point cloud.} Radar data are often represented either as point clouds or as raw multi-dimensional tensors, trading off lower memory and bandwidth requirements in the point-cloud format against richer, more detailed feature information in the tensor representation \cite{caesar2020nuscenes, palffy2022multi, meyer2019automotive}.
Because ADC-sampled radar features encode intuitive physical information, applying a CFAR algorithm \cite{blake1988cfar, farina1986review} isolates large echoes and converts the data into a sparse point-cloud representation.
\cite{kim2024crt} uses camera data to compensate for the information loss from point cloud representation.

\paragraph{Radar compression.}
Prior work has used autoencoders to compress radar data \cite{hu2023radar,xu2021synthetic,stephan2021radar}.
However, the use of autoencoder-based compression would be prohibitive for resource-constrained automotive sensors, as the sensor frontend has limited compute and memory resources. 
Autoencoders used for image compression cannot be deployed on radar frontends since they often have more parameters than the detection or segmentation models themselves.

Recognizing the need to alleviate data transmission overhead, in-sensor radar data compression based on an exponential Golomb encoder is developed in \cite{mani2019memory}; however, the compression ratio is limited to 33\%.
\cite{ding2024radarocc} proposed an energy-based compression scheme using index-value pairs for 3D occupancy prediction. 
While their method leverages spherical encoding and self-attention to compress 4D radar data, it is not compatible with existing neural networks and does not support reconstruction of the original radar tensor.
The reconstructed signal could be useful for downstream tasks, such as motion prediction and planning \cite{weng2024drive}.

\paragraph{Open-loop compression.}
In the image domain, \cite{xie2022bandwidth, xie2019source} optimizes the quantization table from JPEG to improve the rate-distortion trade-offs, whereas \cite{NEURIPS2021_7535bbb9} studies learning-based task-aware compression.
To improve efficiency, others have focused on adaptive spatial bit allocation~\cite{rippel2017real} and modeling latent dependencies via 3D CNNs~\cite{mentzer2018conditional}.
However, since their focus is on improving the rate-distortion trade-off without considering autoencoder complexity, the large model size~\cite{balle2018variational, balle2017endtoend} renders the approach impractical for deployment on the radar sensor side due to energy constraints.
Additionally, these arts employ an \textit{open-loop} approach, focusing on offline compression optimization to enhance task performance under a fixed I/O constraint.
In contrast, we \textit{close} this loop by performing online test-time optimization, adaptively controlling the compression ratio in real time based on task-driven feedback.
    
\begin{figure}[t]
    \centering
    \includegraphics[width=1.0\linewidth]{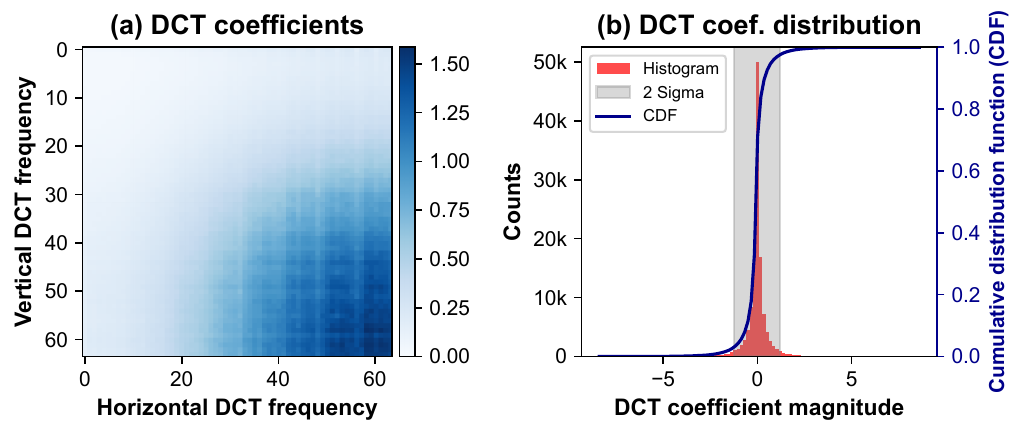}
    \vspace{-1.5em}
    \caption{\textbf{Motivation for spectral pruning and quantization.} (a) The DCT coefficient magnitudes are clustered in the high-frequency bins. (b) Their histogram is sharply peaked,  highlighting strong sparsity and clear opportunities for compression.}
    \label{fig:DCT_coef}
    \vspace{-1em}
\end{figure}

\section{Proposed methods} \label{sec:methods}

\subsection{Spectral compression frontend} \label{ssec:DCT}

\paragraphtight{Overview.}
We compress the raw radar data cube by pruning spectral-domain coefficients.
We find that the energy of radar features is concentrated in a few spectral bins, specifically the high-frequency components, for the RADIal dataset.
The coder operates by dividing features into blocks, applying the DCT, and truncating the coefficients, whereas the decoder restores the data through inverse DCT (IDCT) followed by block merging.
The coder-decoder pair also employs scaled quantization to further reduce data size, while preserving the high dynamic range of radar features.

Figure~\ref{fig:DCT_coef}(a) shows the distribution of the DCT coefficients from the raw radar tensor, computed block‐wise with $M=64$.
Here, we calculate the average coefficient magnitude $\mathbb{E}_{c,b}\big[|\boldsymbol{z}_{c,b}|\big]$ over all channels $c$ and blocks $b$.
This reveals a strong concentration of energy in the high-frequency bins, highlighting the opportunity for aggressive pruning.
Section \ref{sec:dct_remark} includes energy maps of radar features from more radar perception datasets.
Figure~\ref{fig:DCT_coef}(b) presents the histogram of these coefficients, wherein its pronounced peak near zero highlights the potential for quantization.

\paragraph{Problem formulation.}
We first define the radar feature map and then describe adaptive spectral pruning.
The radar generates a complex-valued feature map $\mathcal{X} \in \mathbb{C}^{C\times H\times W}$, which is processed into a real-valued feature map $\mathbf{x} \in \mathbb{R}^{2C \times H \times W}$ by concatenating the real and imaginary parts along the channel axis.
After splitting $\mathbf{x}$ into $M {\times} M$ non-overlapping blocks, we flatten them into $\tilde{\mathbf{x}} \in \mathbb{R}^{2C \times B \times M^2}$ and apply DCT along the last dimension.
This creates DCT coefficients $\boldsymbol{z} = z_{c,b,m}$, where $1 \leq c \leq 2C$ is the channel index, $1 \leq b \leq B = HW/M^2$ is the block index, and $1 \leq m \leq M^2$ is the frequency bin.
We refer readers to Section~\ref{sec:dct} for details on the Type-II DCT.
Returning the coefficients back to the spatial domain involves IDCT for each $M{\times}M$ block.
After that, merging the blocks gives the reconstructed radar feature $\hat{\mathbf{x}} \in \mathbb{R}^{2C \times H \times W}$.

\paragraph{Adaptive spectral pruning.}
Adaptive pruning suppresses smaller spectral coefficients by setting them to zero based on a dynamic pruning ratio.
Given a pruning ratio ${r} \in [r_\text{min},r_\text{max}] =[1,M^2]$, we set index $k = \lfloor M^2/r \rfloor, 1 \leq k \leq M^2$ where $\lfloor \cdot \rfloor$ denotes the floor function.
We write the absolute‐value order‐statistics in descending order $\bigl|z_{c,b,(1)}\bigr| \ge \bigl|z_{c,b,(2)}\bigr| \ge \cdots \ge \bigl|z_{c,b,(M^2)}\bigr|$ so that  the coefficient of $k$-th largest magnitude becomes the blockwise threshold $\,z_{c,b,(k)}=\kappa_{c,b}$.
The pruned coefficients $z'_{c,b,m} = z_{c,b,m} \cdot{} \mathbf{1} \bigl( \lvert z_{c,b,m}\rvert \geq \kappa_{c,b}\bigr)$ now has $k$ non-zero elements where $\mathbf{1}(\cdot)$ denotes an indicator function.

\paragraph{Scaled quantization.}
Our scaled quantization preserves the high dynamic range of radar features.
The first step computes each block’s peak magnitude,
\begin{equation}
    Q_{c,b} \;=\; \max_{1 \le m\le M^2}\,\bigl|z'_{c,b,m}\bigr|.
\label{eq:peak_mag}
\end{equation}
We calculate the per‐block scale factor $\Delta_{c,b} = {Q_{c,b}}/{S}$ based on the maximum positive code $S=2^{s-1}-1$ for an $s$-bit signed uniform quantizer. 
After that, the quantized coefficient $ q_{c,b,m} = \bigl\lfloor {z'_{c,b,m}}/{\Delta_{c,b}} \bigr\rceil \in[-S,S]$ where $\lfloor \cdot \rceil$ denotes rounding to the nearest integer.
The dequantization process reconstructs the DCT coefficients using the scale factor $ \hat z_{c,b,m} = q_{c,b,m}\,\Delta_{c,b}$.
The scale factor only incurs transmission overhead of ${s_\texttt{FP}}/\bigl({s_\texttt{FxP}}{M^2}\bigr)$ where ${s_\texttt{FP}}$ and ${s_\texttt{FxP}}$ respectively indicate floating and fixed point quantization bit widths, e.g., 0.097\% for a 64 \texttimes{} 64 patch with 8-bit quantization.
Larger blocks decrease scale-factor overhead at the cost of patch-level granularity.

\paragraph{Comparison with JPEG.} 
We compare our adaptive codec with JPEG, which employs a fixed quantization table and operates at a fixed compression rate.
The key distinction lies in our approach to compression: we prune by selectively retaining spectral bins based on their magnitudes. 
In contrast, JPEG compression attenuates high-frequency components within each $M{=}8$ block through a fixed quantization table.
We also adopt larger patches (e.g., $M{\in}\{32,64\}$), taking advantage of the inherent sparsity of radar features to amortize the scale-factor overhead from quantization, while also enabling more fine-grained gradient computation.
Ours treats the antenna-pair channels equally, unlike JPEG, which uses RGB-to-YCbCr conversion and chroma subsampling.
These differences stem from the fact that JPEG attempts to preserve information perceivable by the human eye, albeit with compression.
Lastly, JPEG's final stage utilizes lossless bitstream compression, including Huffman and run-length encoding, which are applicable but not employed in our scheme to reduce compression latency.


\subsection{Adaptive compression with proxy gradient} \label{ssec:adaptive}

\begin{figure}[t]
    \centering
    \includegraphics[width=1.0\linewidth]{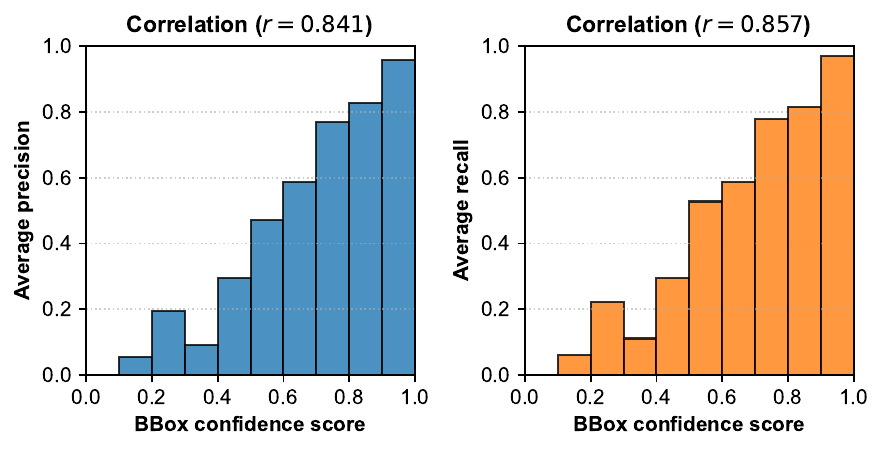}
    \caption{\textbf{Motivation for surrogate objective choice.} (a) AP and (b) AR vs. confidence score yield a high correlation.}
    \label{fig:corr}
    \vspace{-1em}
\end{figure}

\paragraphtight{Overview.}
The next technique is adaptive compression feedback with gradient descent.
We hypothesize that the compressibility of raw features varies across frames, making a fixed compression rate suboptimal and leading to either underutilized bandwidth or performance degradation.
To address this, we design a feedback loop that adjusts the compression ratio online using proxy gradients.
Since ground-truth labels are unavailable at test time after deployment, we estimate model uncertainty from post-softmax confidence scores.

\paragraph{Surrogate objective.} 
Figure~\ref{fig:corr} reveals a strong positive correlation between bounding‑box confidence and object‑detection performance, thereby justifying the use of confidence as a surrogate objective when optimizing detection accuracy.
In the RADIal dataset \cite{rebut2022raw}, precision and recall are evaluated at discrete confidence thresholds from 0.1 to 0.9. 
This thresholding scheme explains the staircase‑like correlation that appears in the results.

\begin{figure}[t]
 \removelatexerror
    \IncMargin{.5em}
    \begin{algorithm}[H]
    
    \SetNlSty{normaltext}{}{:}
    
    \caption{Adaptive compression rate control}
    \label{alg:ada}
    
    \KwIn{Radar sequence $\bigl( \{\mathbf{x}_t\}_{t=1}^T \bigr)$, weight ($\mathbf{\Theta}$), initial pruning ratio ($r_1$), adaptation rate ($\eta$), perturbation ($\epsilon$)}
    \KwOut{Detection map ($\{\hat{\mathbf{y}}_t\}_{t=1}^{T}$)}
    
    \For{$t := 1$ to $T$}
    {
        $\mathbf{q}_t, \mathbf{Q}_t \leftarrow \operatorname{Compression} (\mathbf{x}_t, r_t)$ \tcp*[r]{Transfer $\{\mathbf{q}, \mathbf{Q}\}$ once}
        $\hat{\mathbf{x}}_t \leftarrow \operatorname{Decompression} (\mathbf{q}_t, \mathbf{Q}_t)$\;
        \uIf{adaptive}{
            $\mathbf{q}_t^- \leftarrow \operatorname{Pruning}(\mathbf{q}_t, \epsilon)$\;
            $\hat{\mathbf{x}}_t^- \leftarrow \operatorname{Decompression} (\mathbf{q}_t^-, \mathbf{Q}_t)$\;
            $ \{ \hat{\mathbf{y}}_t, \hat{\mathbf{y}}_t^- \} \leftarrow f_{\mathbf{\Theta}}( \{ \hat{\mathbf{x}}_t, \hat{\mathbf{x}}_t^- \})$ \tcp*[r]{Batch}
            $ \{(b_{k},p_{k})\}_{k} = \operatorname{BBoxPropose}(\{ \hat{\mathbf{y}}_t, \hat{\mathbf{y}}_t^- \})$\;
            \uIf{$k > 0$}
            {
                $r_{t+1} \leftarrow r_t - \eta \hat\nabla_{r_t} J $\; 
            }
        }
        \uElse{
            $\hat{\mathbf{y}}_t \leftarrow f_{\mathbf{\Theta}}( \hat{\mathbf{x}}_t)$\;
            $ \{(b_{k},p_{k})\}_{k} = \operatorname{BBoxPropose}(\hat{\mathbf{y}}_t)$\;
        }
    }
    \end{algorithm}
    \vspace{-1em}
\end{figure}

\paragraph{Task objective.} 
We express the objective as
\begin{equation}
    \label{eq:obj}
    \max_{\{r_t\}_{t=1}^{T}}
    \mathbb{E}\bigl[ J_t(r_t) \bigr],
    \quad
    J_t(r_t)=h(\mathbf{x}_t,r_t)\;-\;\lambda \cdot B(r_t)
\end{equation}
where $\lambda$ controls the accuracy–bandwidth trade‑off and $B(\cdot)$ denotes the instantaneous bit rate, and $h : (\mathbf{x}, r) \mapsto \mathbb{R}$ encapsulates compression, decompression, forward inference, and bounding‑box proposal.
This formulation stems from our hypothesis that a time-varying sequence of pruning ratios $\{r_t\}_{t=1}^T$ can achieve better performance than a single, uniformly applied fixed pruning ratio.
We perform gradient descent (or ascent for minimizing the negative term) by calculating the gradient
\begin{equation}
    \nabla_{r} J = \nabla_{r} h(\mathbf{x}, r)\; - \; \lambda \cdot B'(r).
\end{equation}

\paragraph{Proxy gradient.} 
The proxy gradient calculation is performed entirely on the compute side, eliminating the need for backpropagation and avoiding the transmission of costly backward tensors over the communication link.
Given an input $\hat{\mathbf{x}}$, the network returns a classification map $\hat{\mathbf{y}}_\text{cls} \in [0,1]^{\frac{H}{4}\times\frac{W}{8}}$ and a regression map $\hat{\mathbf{y}}_\text{reg} \in \mathbb{R}^{2 \times \frac{H}{4}\times\frac{W}{8}}$.
We write the combined output as $\hat{\mathbf{y}} = (\hat{\mathbf{y}}_\text{cls}, \hat{\mathbf{y}}_\text{reg})$, where $f_{\mathbf{\Theta}}(\cdot)$ denotes the network function.
Each grid cell is then decoded into a bounding‐box $b\in \mathbb{R}^6$ with a corresponding confidence $p\in[0,1]$ by applying confidence thresholding followed by non‑maximum suppression.
We abstract this operation as $ \{(b_{k},p_{k})\}_{k=1}^{K} = \operatorname{BBoxPropose}(\hat{\mathbf{y}}_\text{cls}, \hat{\mathbf{y}}_\text{reg})$ with $K$ proposals and use the maximum confidence $p_\text{max} = \max_k p_k$.
We perform gradient descent by optimizing $r_t$ sequentially for each time step, and computing the gradient with the negative perturbation as $p^-$ as follows:
\begin{equation}
\hat\nabla_r h(\mathbf{x},r) \approx \frac{h(\mathbf{x},r) -h(\mathbf{x},r-\epsilon)}{\epsilon}=\frac{p-p^-}{\epsilon}.
\end{equation}
In case of a segmentation task, we instead minimize the average entropy $\mathcal{H}(\cdot)$ calculated based on the pixel-wise class probability $\hat{\mathbf{y}}_{\text{seg}} \in [0,1]^{C_s \times H_s \times W_s}$ as $\mathcal{H}( \hat{\mathbf{y}}_{\text{seg}} ) = \mathbb{E}_{(i,j)} [ \mathbb{H} (\hat{\mathbf{y}}_{\text{seg}}^{(i,j)} )]$ where $\mathbb{H}(\cdot)$ is the Shannon entropy.

\paragraph{Adaptive compression rate control.} 
Algorithm~\ref{alg:ada} outlines the adaptive rate control.
Starting with an initial pruning ratio \(r_{1}\), the encoder transforms each feature map $\boldsymbol{}{x}_{t}$ into pruned DCT coefficients $\mathbf{q}_{t}$ and their associated quantization scales $\mathbf{Q}_{t}$ via the DCT, pruning, and quantization stages described in Section~\ref{ssec:DCT}.
At the receiver, dequantization is followed by an IDCT, which reconstructs $\hat{\mathbf{x}}_{t}$; the detector then produces the prediction map $\hat{\mathbf{y}}_{t}$.
Whenever the bit-rate budget must be adjusted, the pruning ratio is updated with a proxy-gradient step: a small perturbation $\epsilon$ is applied to $r_{t}$ (Line~6), additional coefficients are pruned to obtain $\mathbf{q}_{t}^{-}$, and a second inference pass yields \(\hat{\mathbf{y}}_{t}^{-}\).
The difference in bounding-box confidence between \(\hat{\mathbf{y}}_{t}\) and \(\hat{\mathbf{y}}_{t}^{-}\) provides a zeorth-order gradient estimate, which drives a simple gradient–descent update to produce \(r_{t+1}\).
Each adaptation step requires only two forward passes and no backpropagation.

\begin{table*}[t!]
    \caption{\textbf{Compression on RADIal.}
    We compare the detection and segmentation performance of adaptive rate control with prior work on the RADIal \textit{test} set using FFTRadNet as the backbone. Our adaptive rate control achieves 101\texttimes{} compression on average while maintaining the performance close to the baseline. `P' and `R' respectively stand for precision and recall. We highlight in bold the best-performing result among methods other than the baseline.}
    \centering
    \small %
    \begin{tabular}{c|cc|c|ccc|c}
    \toprule
    \multirow{2}{*}{\textbf{Method}} & \multirow{2}{*}{\textbf{Bit}} & \textbf{Prune} & \textbf{Bit rate [bpp]} $\downarrow $ & 
    \multicolumn{3}{c|}{\textbf{Detection}} & \textbf{Segmentation} \\
                                     &                               & \textbf{ratio} $\uparrow$ & \textbf{(Comp. ratio $\uparrow$)} & \textbf{P} (\%) $\uparrow$ & \textbf{R} (\%) $\uparrow $& \textbf{F\textsubscript{1}} (\%) $\uparrow$ & \textbf{mIoU} (\%) $\uparrow$ \\
    \midrule
    Uncompressed baseline \cite{rebut2022raw}        & 32 & - & 32 (1\texttimes{})                                                   & {97.24}     & {95.93}  & {96.58} & {75.97} \\
    \midrule
    Index-value \cite{ding2024radarocc}             & 32 & 12\texttimes{} & {2.67} ({12\texttimes{}})                               & \textbf{97.55} & {62.12} & {75.91} & {49.86} \\
    Adaptive compression (Ours)                     & 4 & 12.57\texttimes{} & \textbf{0.32} (\textbf{101\texttimes{}})                   & {96.25} & \textbf{94.04} & \textbf{95.13} & \textbf{79.34} \\
    \bottomrule
    \end{tabular}
    \label{tab:radial}
\end{table*}

\section{Experiments} \label{sec:experiments}

\subsection{Experimental setup} \label{ssec:exp_set}

\paragraphtight{RADIal dataset.} 
We use the RADIal \cite{rebut2022raw} dataset to demonstrate the efficacy of the proposed method.
It consists of 8,252 labelled scenes, each consisting of a camera image, raw radar measurements, and laser scans.
The post-FFT radar feature tensor has dimensions of 32 channels, 512 range bins, and 256 Doppler bins, with each element represented as a 32-bit floating-point value. 
We adhere to the three-fold dataset split (training, validation, and test) provided with the dataset in \textit{sequence} mode.

\paragraph{CARRADA dataset.}
We also validate our results on the CARRADA dataset \cite{ouaknine2021carrada}, which also provides camera data.
Measured on 76-to-81GHz mmWave radar \cite{instruments2018single}, this dataset is designed to serve as a benchmark for radar-based semantic segmentation.
It comprises 30 sequences totaling 7,193 annotated frames, providing object-level segmentation labels for pedestrians, cars, and cyclists.
Each radar frame is processed into a post-FFT tensor with 256 range bins and 64 Doppler bins, encoded as 32-bit tensors.
We follow the prescribed training/validation/test split routines and use the \textit{sequence} configuration for online adaptation.

\paragraph{Radatron dataset.}
In addition, we evaluate our approach on the Radatron dataset \cite{radatron}, which captures diverse urban driving scenarios with synchronized high-resolution radar and camera data.
The dataset comprises approximately 16,000 annotated frames, each labeled with vehicle bounding boxes to support object detection tasks.
Measured with a cascaded MIMO radar \cite{TI_IWR1843_2024}, each radar frame is processed into a range-azimuth heatmap with 448 range bins and 192 azimuth bins, represented as 32-bit data.
We adopt the official training/validation/test split and operate in the \textit{sequence} configuration to support online rate adaptation at test time.

\paragraph{Implementation details.}
We train the FFTRadNet model \cite{rebut2022raw} from scratch by using PyTorch \cite{paszke2019pytorch} on a single NVIDIA Quadro RTX 6000 for the evaluation on the RADIal dataset.
We apply on-the-fly compression and quantization to the raw radar features by sampling the pruning ratio $r \sim U(r_{\min},r_{\max})$ and quantization bit width $s_\texttt{FxP}$-bit.
We train with a batch size of $4$ and up to $100$ epochs.
 The Adam optimizer \cite{kingma2015adam} was used with a learning rate of $10^{-4}$, decaying by 0.9 every 10 epochs.

\paragraph{Metric.} 
For the RADIal dataset \cite{rebut2022raw}, average precision (AP) and average recall (AR) are calculated by averaging precision and recall measurements at thresholds $\{0.1, 0.2,...,0.9\}$.
The F\textsubscript{1} score is the harmonic mean of AP and AR.
For the Radatron dataset \cite{radatron}, we report average precision at IoU thresholds of 0.5 (AP\textsubscript{50}) and 0.75 (AP\textsubscript{75}) as primary evaluation metrics. 
We also compute the mean Average Precision (mAP) averaged over IoU thresholds ranging from 0.5 to 0.95 in increments of 0.05.
On the other hand, for the segmentation tasks, mean intersection over union (mIoU) is measured by comparing the predicted drivable freeway map with the target map.
In addition to the IoU metric, we use the dice score for evaluation with the CARRADA dataset \cite{ouaknine2021carrada}.

\subsection{Experimental results} \label{ssec:main_res}

\paragraphtight{Adaptive compression.}
Table \ref{tab:radial} compares our adaptive compression method with the prior index-value-based method on the RADIal dataset.
Ours achieves a twofold improvement in the compression ratio while maintaining an F\textsubscript{1} score drop of less than 1.5\%p.
An average pruning ratio of 12.57\texttimes{} combined with 8\texttimes{} gain from quantization, yields a bit rate of 0.32 bits per pixel (bpp), corresponding to a 101\texttimes{} compression ratio.
On the segmentation task, ours actually improves the performance by 4\%.
Here, we report the compression ratio with respect to the number of non-zero elements for both methods.
Although the index-value pair-based method offers slight improvements in the precision, it significantly degrades both detection and segmentation performance.
We report ablation studies in Sec. \ref{sup:ablation} of the Supplementary Material.

\begin{table*}[t]
    \caption{\textbf{Compression on CARRADA.} 
    We compare our adaptive compression strategy with prior arts on mIoU and mDice metrics on the CARRADA dataset using the TMVA-Net. Our adaptive method maintains segmentation performance close to the baseline despite a large reduction (117\texttimes{}) in data rate. `PR', `BR', and `CR' respectively stand for the prune ratio, bit rate, and total compression ratio. Bold text indicates the top-performing results, excluding the baseline.}
    \centering
    \small
    \begin{tabular}{c|cc|cc|c|c}
    \toprule
    \textbf{Method} & \textbf{Bit} & \textbf{PR} $\uparrow$ & \textbf{BR [bpp]} $\downarrow$ & \textbf{CR} $\uparrow$ & \textbf{mIoU} (\%) $\uparrow$ & \textbf{mDice} (\%) $\uparrow$ \\
    \midrule
    \multirow{1}{*}{Uncompressed baseline \cite{ouaknine2021multi}} & 32 & - & 32 & 1\texttimes{} & {55.25} & {67.13} \\
    \midrule
    Index-value \cite{ding2024radarocc}     & 32  & 29\texttimes{}  & 1.10 & 29\texttimes{}      & 38.96 & 46.90\\
    Adaptive compression (Ours)    & 8  & 29.26\texttimes{} & \textbf{0.27} & \textbf{117\texttimes{}}     & \textbf{54.03} & \textbf{65.87} \\
    \bottomrule
    \end{tabular}
    \label{tab:carrada}
\end{table*}

\begin{table*}[t]
    \caption{\textbf{Compression on Radatron.} 
    We compare compression strategies on mAP, AP$_{50}$, and AP$_{75}$ while reducing the bit rate. Our adaptive method achieves higher detection performance than both the baseline and index-value compression while operating at a much lower bit rate (30$\times$ compression). `PR', `BR', and `CR' respectively stand for the prune ratio, bit rate, and total compression ratio. The best-performing result, except for the baseline method, is highlighted in bold.}
    \centering
    \small
    \begin{tabular}{c|cc|cc|c|c|c}
    \toprule
    \textbf{Method} & \textbf{Bit} & \textbf{PR} $\uparrow$ & \textbf{BR [bpp]} $\downarrow$ & \textbf{CR} $\uparrow$ & \textbf{mAP} (\%) $\uparrow$ & \textbf{AP\textsubscript{50}} (\%) $\uparrow$ & \textbf{AP\textsubscript{75}} (\%) $\uparrow$ \\
    \midrule
    Baseline \cite{radatron} & 32 & - & 32 & 1\texttimes{} & {46.07} & {83.60} & {44.16} \\
    \midrule
    +Index-value \cite{ding2024radarocc} & 32 & 7.5\texttimes{} & 4.27 & 7.5\texttimes{} & 45.72 & 80.44 & 47.54 \\
    +Adaptive (Ours) & 8 & 7.5\texttimes{} & \textbf{1.07} & \textbf{30\texttimes{}} & \textbf{48.46} & \textbf{83.69} & \textbf{49.07} \\
    \bottomrule
    \end{tabular}
    \label{tab:radatron}
    \vspace{-1em}
\end{table*}

In Table \ref{tab:carrada}, we demonstrate the generality of our compression method even on the segmentation task by using the CARRADA dataset.
It achieves an even greater compression ratio of 117\texttimes{} while maintaining the target metric close to the non-compressed baseline, with \textasciitilde{}1\% drop.
On the other hand, the prior art \cite{ding2024radarocc} suffers from significant degradation in segmentation performance.

We also validate the efficacy of our online compression method on the Radatron dataset and present it in Table \ref{tab:radatron}.
Although the radatron dataset only provides a range-azimuth heatmap, it is still compressible by 30\texttimes{} with our compression method.
Even at this compression rate, ours outperforms the non-compressed baseline by up to 5\% in AP\textsubscript{75}.
We hypothesize that feature compression works as a filter, helping to suppress noise and clutter. 
While the prior method improves AP\textsubscript{75}, it lowers performance in others.

\begin{figure}[t]
    \centering
    \includegraphics[width=0.95\linewidth]{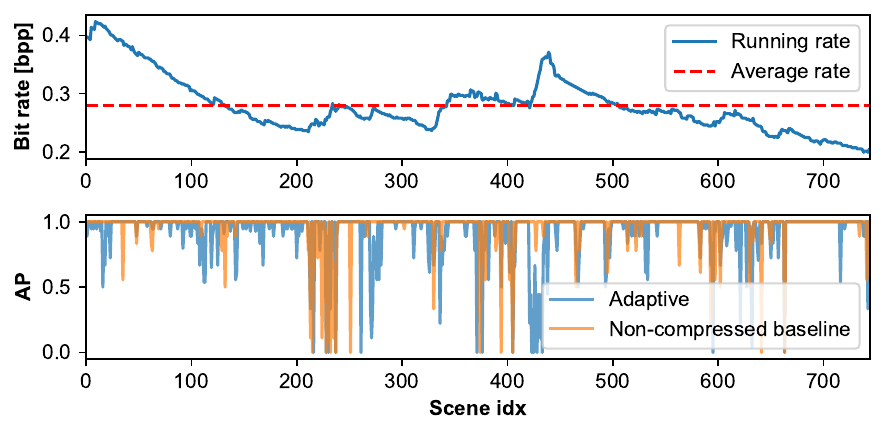}
    \vspace{-1em}
    \caption{\textbf{Online adaptation.} 
    Bit rate time series (above). Scene-wise average recall change (below).
    }
    \label{fig:time_series}
    \vspace{-1em}
\end{figure}

\paragraph{Online adaptation.}
The online adaptation adjusts the bit rate based on the confidence gradient.
Figure~\ref{fig:time_series} illustrates this scheme, where the pruning ratio is increased from the initial starting point.
The bit rate fluctuates over time, though it surges at some points.
Here, it is expected that the controller decreases the pruning ratio to compensate for the AR drop.
It achieves an average bit rate of {0.279} bpp with 8-bit quantization, yielding a 115\texttimes{} compression with AR of {93.91\%}.
We validate the robustness of our approach under a noisy scenario and report the adversarial results in Table \ref{tab:stability} of the Supplementary Material.

\begin{figure}[t]
    \vspace{-0.5em}
    \centering
    \includegraphics[width=1.0\linewidth]{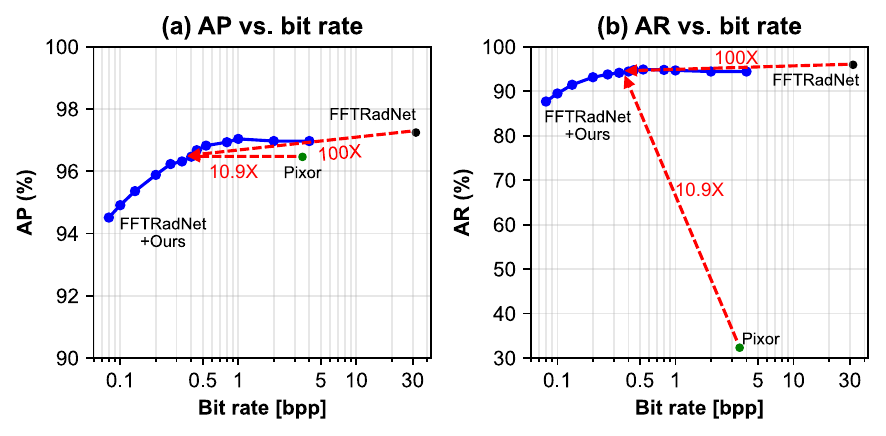}
    \vspace{-1.5em}
    \caption{\textbf{Rate-accuracy tradeoff.} 
    (a) Average precision and (b) average recall vs. bit rate measured with a static compression rate on the RADIal test set with $4$-bit quantization.
    }
    \vspace{-1em}
    \label{fig:RA_trade}
\end{figure}

\paragraph{Rate-accuracy tradeoff.}
Figure~\ref{fig:RA_trade} illustrates the trade-off between the bit rate and the detection performance metrics: AP and AR.
While it depicts the obvious rate-accuracy tradeoff, our compression scheme achieves a 100\texttimes{} reduction in the radar feature map size while incurring a \textasciitilde{}1\%p decrease in performance compared to FFTRadNet \cite{rebut2022raw}.
Compared with Pixor \cite{yang2018pixor}, our method achieves higher precision and markedly higher recall.
Here, we use the network retrained on radar features compressed and decompressed using the setup explained in Sec. \ref{ssec:exp_set}.
Figure~\ref{fig:RA_trade} shows that accuracy remains virtually unchanged down to 1 bpp.
This enables us to set the lower bound on the pruning ratio, $r_\text{min}$, at this point for adaptive control, thereby reducing the minimum channel bandwidth requirement.
In particular, $r_\text{min}$ determines the peak bit‑rate requirement of the sensor‑to‑compute link.

\begin{figure*}[t]
	\centering
	\includegraphics[width=0.95\textwidth]{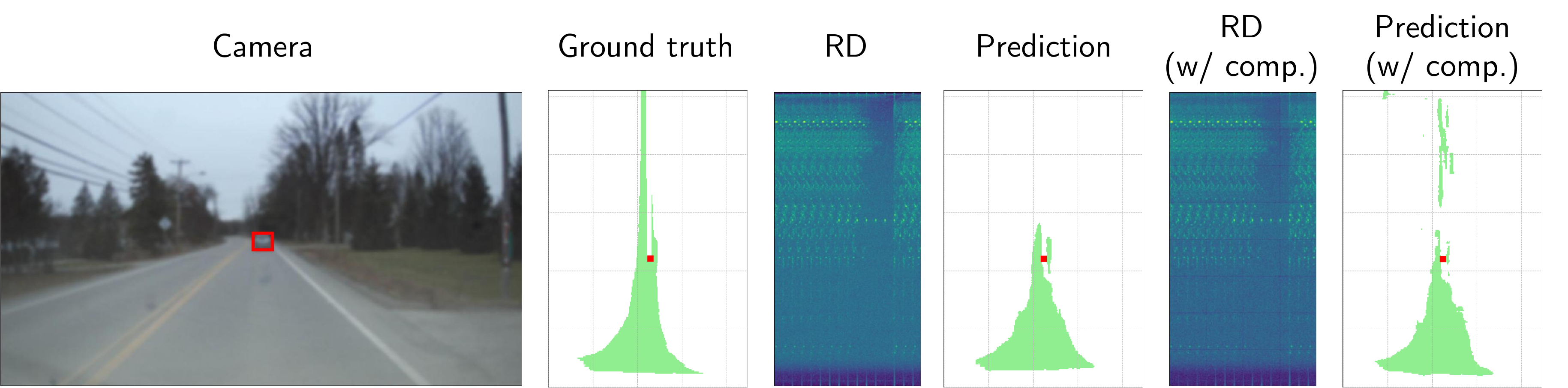}
    \caption{\textbf{Qualitative comparisons.} 
    Networks are trained on raw radar tensors, visualized as range-Doppler (RD) images, along with ground-truth labels for freeway segmentation and vehicle detection. Camera images only provide contextual reference for the scene. RD images map Range to the $y$-axis and Doppler to the $x$-axis. We highlight the detected car in {\color{red} red} and the segmented map in {\color{green} green} in the bird’s-eye-view visualization. Notably, the compressed RD outperforms the uncompressed baseline.}
    \vspace{-1em}
	\label{fig:qual}
\end{figure*}

\paragraph{Qualitative results.} 
The qualitative comparison between a non-compressed and compressed radar tensor is shown in Fig. \ref{fig:qual}.
There is no distinction between the range-Doppler (RD) image against the compressed counterpart as depicted in the drawing.
Interestingly, the network output with compressed features yields better performance than the non-compressed counterpart.
Here, a pruning ratio of 5 is used with no quantization.
We surmise that spectral pruning improves downstream performance by filtering out noise and clutter picked up by the radar.

\begin{figure}[t] 
    \centering
    \includegraphics[width=0.75\linewidth]{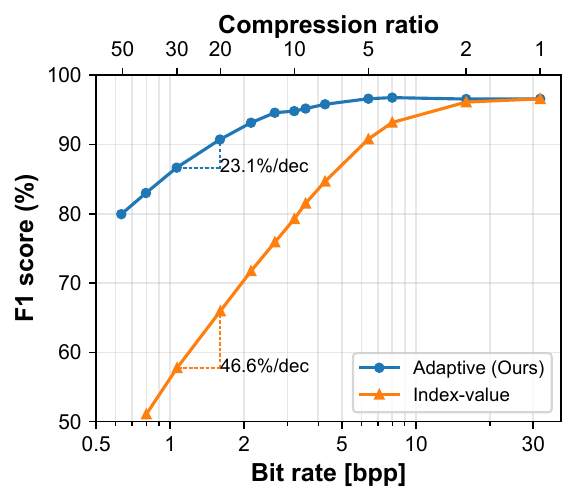}
    \vspace{-1em}
    \caption{\textbf{Adaptive spectral pruning vs. index-value-based compression.} Adaptive spectral pruning-based compression outperforms index-value-based compression \cite{ding2024radarocc} with much more resiliency to error. Both methods utilize \textit{Float32}, and are tested on the RADIal test set.}
    \label{fig:vs_index_value}
    \vspace{-1em}
\end{figure}

\paragraph{Adaptive spectral pruning vs. index-value-based compression.} Figure~\ref{fig:vs_index_value} demonstrates how the adaptive spectral pruning-based compression method outperforms the index-value pair-based method \cite{ding2024radarocc}.
While the performance drop-off begins immediately with the index-value pair-based method, our method remains stable until 5\texttimes{} compression.
On top of that, the roll-off gradient of 23.1\%/dec is much gentler for the spectral pruning-based compression than the 46.6\%/dec of the counterpart.
The index-value-based compression selects the Top-$K$ elements in each $M {\times} M$ block based on energy in the spatial domain.
Spectral-domain compression retains much more information than spatial-domain compression.
Both methods use the 32-bit floating-point (\textit{Float32}) representation, and the baseline model is used without any fine-tuning.

\section{Conclusion} \label{sec:conclusion}

\noindent We present \textbf{AdaRadar}, a rate-adaptive spectral-domain compression framework for radar-based object detection.
Pruning and quantizing spectral coefficients shrink the range–Doppler cube by over 100\texttimes{}, easing the sensor‑to‑compute bandwidth bottleneck while preserving accuracy at the lowest bit rate target.
A lightweight test-time controller finely adjusts the compression ratio on the fly, making a judicious tradeoff between bandwidth and performance in real time.
Experiments on the RADIal, CARRADA, and Radatron datasets demonstrate the robustness of our compression approach.

\paragraph{Limitations.}
Although heavily explored in the literature, the compression of camera images is not considered.
The adaptive compression might fall short if the temporal relationship among subsequent frames is not guaranteed.

\section*{Acknowledgements} 
\noindent This work was supported in part by COGNISENSE, one of seven centers in JUMP 2.0, a Semiconductor Research Corporation (SRC) program sponsored by DARPA. 
The work of SY Chun was supported by IITP grants funded by the Korea government(MSIT) [No.RS-2021-II211343, Artificial Intelligence Graduate School Program (Seoul National University) / No.RS-2025-02314125, Effective Human-Machine Teaming With Multimodal Hazy Oracle Models].

{
    \small
    \bibliographystyle{unsrtnat} 
    \bibliography{main, ulvio}
}

\clearpage

\maketitlesupplementary

\setcounter{section}{0}
\setcounter{figure}{0}
\setcounter{table}{0}
\setcounter{equation}{0}

\renewcommand{\thesection}{S\arabic{section}}
\renewcommand{\thefigure}{S\arabic{figure}}
\renewcommand{\thetable}{S\arabic{table}}
\renewcommand{\theequation}{S\arabic{equation}}

\noindent The Supplementary Material is organized as follows:
\begin{itemize}
    \item Section~\ref{sup:exp_setup} provides implementation details, including network architectures, training protocols, hardware resources, and asset licenses.
    \item Section~\ref{sup:add_note} elaborates on notations for transform and alternative task objectives.
    \item Section~\ref{sup:quant_results_be} presents extended quantitative evaluations on the adaptive control, covering ablation studies on hyperparameters and surrogate objective, and stability analysis.
    \item Section~\ref{sup:quant_results_fe} presents additional quantitative evaluations on the spectral compression, covering quantization and block length modulation.
    \item Section~\ref{sec:ext_base} presents additional baselines, including compression using autoencoders, H.264, Gumbel-sigmoid method, and CFAR algorithm.
    \item Section~\ref{sup:qual_results} illustrates the qualitative impact of AdaRadar through range-Doppler maps and visual performance comparisons.
\end{itemize}

\section{Experiment setup details}\label{sup:exp_setup}

\subsection{Baseline specifications and architectures.}

\noindent The baselines in Table \ref{tab:radial}--\ref{tab:radatron}, refer to the corresponding vanilla models without any compression. 
We demonstrate AdaRadar's compression capability across three datasets with distinct model architectures to showcase the \textit{robustness} of our approach.
\begin{itemize}
    \item The baseline, FFTRadNet \cite{rebut2022raw} (Table \ref{tab:radial}), detects vehicles and free driving spaces by using a learned MIMO pre-encoder, a feature pyramidal network (FPN) encoder, and separate detection and segmentation heads.
    \item TMVA-Net \cite{ouaknine2021carrada} in Table \ref{tab:carrada} performs four-class radar semantic segmentation with pedestrian, cyclist, car, and background. It first reshapes the range-azimuth-Doppler (RAD) tensor into azimuth-Doppler (AD), range-Doppler (RD), and range-azimuth (RA) views, then applies 3D convolutions within view-specific encoders.
    \item Radatron \cite{radatron} (Table \ref{tab:radatron}) also runs parallel FPN branches on high- and low-resolution radar streams, which are fused later to output vehicle detection maps.
\end{itemize}

\subsection{Experimental details}

\noindent For Table \ref{tab:radial}, we perform adaptation with a task objective same as Eq. \ref{eq:obj} with $r_\text{initial}=12$, $\eta=1.0$, $\epsilon=0.05$, $p_\text{threshold}=0.8$, $\lambda = 1.0$, $\nabla_\text{clip} = 1.0$.

For Figure \ref{fig:time_series}, we experiment with a task objective same as Eq. \ref{eq:alt_obj} with $r_\text{initial}=20$, $\eta=1.0$, $\epsilon=0.05$, $p_\text{threshold}=0.9$, $\lambda = 15.0$, $\nabla_\text{clip} = 1.0$.

\subsection{Additional training details and resources}

\paragraph{Training details on TMVA-Net.}
All experiments are implemented in PyTorch and trained on a single Nvidia RTX A6000 GPU.
We follow the setting as per the original paper with the Adam optimiser\cite{kingma2015adam}, using its default hyper-parameters: $\beta_{1}=0.9$, $\beta_{2}=0.999$, and $\varepsilon=10^{-8}$).
Training proceeds for 300 epochs with a mini-batch size of 6.
The initial learning rate is set to $10^{-4}$ and decays by a factor of 0.1 every 20 epochs via a StepLR scheduler.

\paragraph{Training details on Radatron.}
We abide by the original training recipe and train the model using stochastic gradient descent (SGD) with a base learning rate of 0.01 on a single Nvidia RTX A6000 GPU.
The learning rate decays by a factor of 0.2 at 15k and 20k iterations, following a step schedule.
Training runs for 25k iterations in total, with a mini-batch size of 8 images.
All other training hyperparameters follow the default settings in Detectron2.

\paragraph{Computation resources.}
The experiments on the RADIal dataset are run on a Red Hat Enterprise Linux 8 workstation equipped with a single 3.5 GHz Intel Core i9-9920X processor, four NVIDIA RTX 6000 GPUs (24 GB VRAM each), 128 GB of system RAM.
The rest of our experiments are conducted on a Red Hat Enterprise Linux 8 server equipped with dual Nvidia RTX A6000 GPUs, each equipped with 48GB VRAM, an AMD Ryzen Threadripper PRO 3995WX CPU @ 4.2GHz, and 256GB of RAM.

\subsection{Assets}

\paragraph{License for RADIal.}
The RADIal dataset\footnote{\url{https://github.com/valeoai/RADIal}} \cite{rebut2022raw} is released without an explicit license; we cite the source and use it solely for non-commercial academic research, in accordance with standard scholarly practices.

\paragraph{License for CARRADA.}
The CARRADA dataset \cite{ouaknine2021carrada} is published under the CC BY-NC-SA 4.0 License, and all CARRADA code\footnote{\url{https://github.com/valeoai/carrada_dataset}} is released under the GNU General Public License v3.0 (GPL-3.0).

\paragraph{License for Radatron.}
The Radatron dataset\footnote{\url{https://github.com/waleedillini/radatronDataset}} \cite{radatron} is released under the Apache License 2.0.

\subsection{Discussions}

\paragraph{Broader impacts.}
By reducing data transmission and memory access, our method could reduce energy usage, contributing to sustainability in high-throughput ML systems.
Radar-based systems improve robustness under adverse weather or poor lighting conditions, potentially reducing accidents in autonomous and assisted driving.

\section{Additional notes on proposed method} \label{sup:add_note}

\subsection{Discrete cosine transform} \label{sec:dct}
\noindent Type-II DCT computation is described by 
$ {\boldsymbol{z}}_{c,b} = \mathbf a \odot \mathbf G \tilde{\mathbf{x}}_{c,b} $ 
where
$\tilde{\mathbf{x}}_{c,b} \in \mathbb R^{M^2}$
is a flattened feature map, $\odot$ denotes Hadamaard (element-wise) product, $\mathbf G \in \mathbb R^{M^2 \times M^2}$ describes cosine coefficients, and $\mathbf a \in \mathbb R^{M^2}$ is a normalization vector.
$\mathbf G$ is computed by $G_{Mu+v,Mi+j} = \cos\bigl(\frac{(2i+1)u\pi}{2M}\bigr) \cos\bigl(\frac{(2j+1)v\pi}{2M}\bigr)$ with indices $i,j,u,v \in \{0,\dots,M-1\}$.
$\mathbf a$ has entries
\begin{equation}
a_{Mu+v} =\frac{2}{M}\,\alpha(u)\,\alpha(v)
\end{equation}
\label{eq:norm_vec}

\begin{equation}
\alpha(u)=
    \begin{cases}
    1/\sqrt2    &   u=0 \\
    1           &   \text{otherwise}
    \end{cases}
\end{equation}
\label{eq:norm_vec2}

\begin{equation}
u,v \in \{0,\dots,M-1\}.
\end{equation}
\label{eq:norm_vec3}

\subsection{Alternative task objective} \label{sec:alt_obj}

\noindent Beyond the objective in Eq. \ref{eq:obj}, we define a constraint-aware gradient
\begin{equation}
    \label{eq:alt_obj}
    \nabla_r J = (p - p_{\min}) + \hat\nabla_r h \cdot (r - r_{\min} + \lambda).
\end{equation}
In this formulation, $p_{\min}$ and $r_{\min}$ denote the lower bounds for confidence and pruning ratio, respectively, while $\hat{\nabla}_r h$ represents the zeroth-order gradient approximation.
The hyperparameter $\lambda$ regulates the Pareto optimal trade-off between task accuracy and transmission bandwidth.

We employ confidence thresholding and gradient clipping to facilitate more stable adaptation.
Specifically, gradient clipping constrains the estimated gradient $\hat\nabla_r h(\mathbf{x},r)$ within a predefined range $[\nabla_{\text{min}}, \nabla_{\text{max}}]$, while confidence thresholding skips the adaptation process for the given timestep if $p < p_\text{threshold}$.

\section{Extended quantitative analysis on adaptive control back end}\label{sup:quant_results_be}

\subsection{Compression/decompression latency}

\begin{figure}[h] 
	\centering
	\includegraphics[width=1\linewidth]{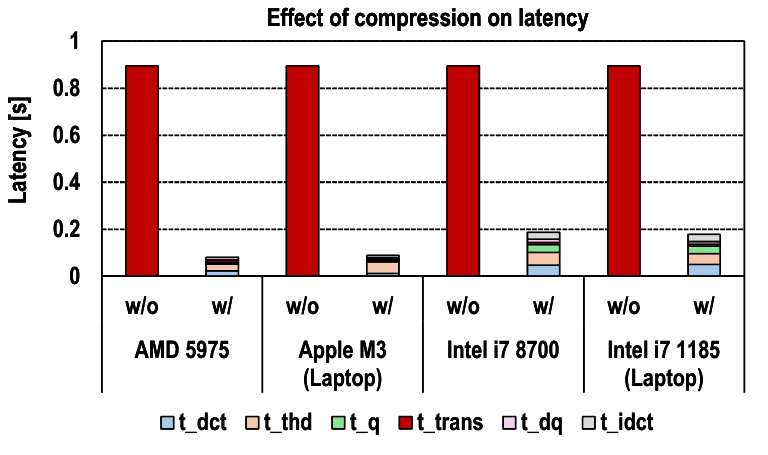}
        \caption{
        \textbf{Compression, decompression, communication cost analysis.} 
        Our compression and decompression processes significantly reduce latency compared to the baseline system, which transmits the original radar tensor.
        }
	\label{fig:latency}
\end{figure}

\noindent In Figure~\ref{fig:latency}, we observe that compression and decompression alleviate the latency required to transfer radar data. 
This is because sensor-to-compute links, such as a CAN bus in automotive systems, have limited bandwidth for transferring high-fidelity data.
Our method can easily re-utilize the embedded DSP block, which is used to compute the fast Fourier transform along range and Doppler dimensions in MIMO radars \cite{TI_IWR1843_2024}.
The radar side only requires simple operations (e.g., DCT, sorting, and rounding) to support AdaRadar.

The total latency arises from three components: (1) compression, (2) transmission, and (3) decompression. More concretely,
\begin{equation}
    t_\text{overall} = \underbrace{t_\text{dct} + t_\text{thd} + t_\text{q}}_{\text{coder}}
                        + t_\text{trans}
                        + \underbrace{t_\text{dq} + t_\text{idct}}_{\text{decoder}}.
\end{equation}
In this experiment, compression and decompression latencies are measured on machines with different CPUs by compressing and decompressing 200 radar tensors of shape 512\texttimes{}256\texttimes{}32 from the RADIal test set.
Transfer time is estimated by assuming a radar \cite{TI_IWR1843_2024} followed by a serial link based on MIPI CSI2 with 150 Mbps bandwidth, sending the radar tensor in a burst mode.

This is analogous to loading a RAW image versus a JPEG-compressed image.
Loading a RAW format takes longer because the loading time is bottlenecked by data \textit{transmission} rather than by decompression.
A typical JPEG decoder \cite{libjpeg} takes about one millisecond to decompress a 512\texttimes{}512 image, while loading the uncompressed image from a CAN bus \cite{TI_TCAN3414_2023} with a bandwidth of a few MB/s would take dozens of milliseconds.

\subsection{Ablation study} \label{sup:ablation}

\paragraph{Surrogate objective ablation.} \label{sec:sur_obj_abl}
We use bounding box confidence as a proxy for task performance, assuming that regions with high detector confidence should be preserved more faithfully. 
To justify our choice, we show additional empirical analysis showing how confidence best correlates with the task performance under compression, and (2) add ablation studies where we replace confidence with alternative signals such as range and azimuth location variances among bounding box predictions per object.

\begin{table}[h]
    \centering
    \small
    \caption{
    \textbf{Correlation between OD prediction output and performance metrics.}
    The performance metrics, such as AP and AR, are highly correlated with confidence but not with other objectives.
    }
    \label{tab:correlation}
    \begin{tabular}{ccc}
    \toprule
    \textbf{Surrogate objective} & \textbf{Corr. w/ AP} & \textbf{Corr. w/ AR} \\
    \midrule
    Confidence (Ours) & \textbf{0.841} & \textbf{0.857} \\
    Range stdev.      & -0.129          & -0.045          \\
    Azimuth stdev.    & -0.070          & -0.016          \\
    \bottomrule
    \end{tabular}
    \label{tab:corr}
\end{table}

In Table \ref{tab:corr}, we demonstrate that the bounding box confidence correlates most closely with the average precision (AP) and average recall (AR). 
On the other hand, the range and azimuth exhibit a weak (negative) correlation. 
We take the standard deviation of the range and azimuth values of predicted bounding boxes.

\begin{table*}[h]
    \centering
    \small
    \caption{
    \textbf{Surrogate objective ablation with range standard deviation, azimuth standard deviation, and a random variable.} 
    We observe that different choices of surrogate objective do not correlate well with the downstream task performance, leading to severe degradation in the perception performance. In those cases, such a significant drop in F\textsubscript{1} score bears similar performance to the random variable.
    }
    \begin{tabular}{c|cc|ccc|c}
    \toprule
    \textbf{Surrogate objective} & \textbf{CR}~$\uparrow$ & \textbf{CR stdev.}~$\downarrow$ & \textbf{Precision}~$\uparrow$ & \textbf{Recall}~$\uparrow$ & \textbf{F\textsubscript{1}}~$\uparrow$ & \textbf{mIoU}~$\uparrow$ \\
    \midrule
    Confidence (Ours) & 6.58 & \textbf{0.91} & \textbf{0.969} & \textbf{0.948} & \textbf{0.958} & \textbf{0.796} \\
    Range stdev.      & 60.9 & 22.1          & 0.941          & 0.785          & 0.856          & 0.721          \\
    Azimuth stdev.    & \textbf{87.0} & 9.8  & 0.930          & 0.762          & 0.837          & 0.708          \\
    Random            & 34.1 & 29.7          & 0.945          & 0.834          & 0.886          & 0.741          \\
    \bottomrule
    \end{tabular}
    \label{tab:alb_obj}
\end{table*}

In Table \ref{tab:alb_obj}, we ablate the surrogate objective with range and azimuth deviations and a random variable $p \sim U(0,1)$, which are fed into our zeroth-order optimizer. 
We find that the compression ratio fluctuates widely, unlike the confidence metric, whose standard deviation across rate sequences remains less than one. This also results in severe performance degradation.

\begin{table*}[h]
    \caption{\textbf{Hyperparameter ablation with $\lambda$.} We analyze the trade-off between bit rate, compression ratio, and downstream performance on detection and segmentation through modulation of $\lambda$. `PR', `BR', and `CR' stand for prune ratio, bit rate, and compression ratio, respectively. We use 4-bit quantization with a block length of $M = 64$. The pruning ratio is initially set to 12, and the learning rate is configured as $\eta = 1$.}
    \centering
    \small
    \begin{tabular}{c|c|c|c|c|c|c|c}
    \toprule
    $\lambda$ & \textbf{PR} $\uparrow$ & \textbf{BR [bpp]} $\downarrow$ & \textbf{CR} $\uparrow$ & \textbf{Precision} (\%) $\uparrow$ & \textbf{Recall} (\%) $\uparrow$ & \textbf{F\textsubscript{1}} (\%) $\uparrow$ & \textbf{mIoU} (\%) $\uparrow$ \\
    \midrule
    1   & 12.57 & 0.32 & 100.56 & 96.25 & 94.03 & 95.13 & 79.34 \\
    3.3 & 13.83 & 0.29 & 110.68 & 96.21 & 93.87 & 95.03 & 79.21 \\
    10  & 16.09 & 0.25 & 128.74 & 96.07 & 93.64 & 94.84 & 79.12 \\
    33  & 20.75 & 0.19 & 166.03 & 95.70 & 93.07 & 94.36 & 78.87 \\
    100 & 27.91 & 0.14 & 223.29 & 95.48 & 91.66 & 93.53 & 78.18 \\
    \bottomrule
    \end{tabular}
    \label{tab:lambda_sweep}
    \vspace{-1em}
\end{table*}

\paragraph{Hyperparameter ablation.} \label{sec:hyp_par_abl}
We vary the regularization weight $\lambda$ to explore how it governs the trade-off between compression efficiency and downstream performance in detection and segmentation.
The experiments are conducted on the RADIal dataset using the FFTRadNet model (Table~\ref{tab:lambda_sweep}).
We apply 4-bit quantization with a fixed block length of $M = 64$. The pruning ratio is initialized to 12, and the learning rate is set to $\eta = 1$.
As $\lambda$ increases, the system adjusts the pruning strength, balancing lower bit rates with the preservation of task-relevant features.
Each configuration corresponds to a Pareto-optimal point that jointly optimizes compression and task performance.

\begin{table*}[h]
    \centering
    \small
    \caption{\textbf{Proxy gradient stability at varying additive Gaussian noise levels on the RADIal test set.}
    We test the stability of our feedback loop by adding Gaussian noise to the radar input (RD map). Even under the noisy scenario, our adaptive control-based compression demonstrates similar performance to the non-compressed baseline.
    }
    \begin{tabular}{c|cc|cc|ccc|c}
    \toprule
    \textbf{Compression} & \textbf{Condition} & \textbf{Noise Scale} & \textbf{Comp. Ratio}~$\uparrow$ & \textbf{Prun. Ratio}~$\uparrow$ & \textbf{Precision}~$\uparrow$ & \textbf{Recall}~$\uparrow$ & \textbf{F\textsubscript{1}}~$\uparrow$ & \textbf{mIoU}~$\uparrow$ \\
    \midrule
    \xmark{} No  & Mild      & 1 & 1.0  & 1.0  & 0.973 & 0.922 & 0.947 & 0.774 \\
    \cmark{} Yes & Mild      & 1 & 36.8 & 4.6  & 0.969 & 0.910 & 0.939 & 0.777 \\
    \xmark{} No  & Moderate  & 2 & 1.0  & 1.0  & 0.957 & 0.884 & 0.919 & 0.694 \\
    \cmark{} Yes & Moderate  & 2 & 33.7 & 4.2  & 0.967 & 0.860 & 0.910 & 0.692 \\
    \xmark{} No  & Severe    & 5 & 1.0  & 1.0  & 0.931 & 0.771 & 0.844 & 0.618 \\
    \cmark{} Yes & Severe    & 5 & 25.6 & 3.2  & 0.939 & 0.720 & 0.815 & 0.602 \\
    \bottomrule
    \end{tabular}
    \label{tab:stability}
\end{table*}

\paragraph{Proxy gradient stability.} 
Extreme fluctuation scenarios induced by high-speed motion or severe occlusion are unfortunately not present in the open-sourced datasets. 
We also believe that occlusion is an unrealistic scenario for radars: (1) a car would be almost crashing into another for a radar having a wide azimuth field of view (FoV) of 140°, and (2) a radar is unaffected by rain or fog, unlike a camera.

We therefore synthetically generate such extreme fluctuation scenarios by adding white noise to the radar tensor and present the results in Table \ref{tab:stability}. 
We mimic scenarios with mild, moderate, and severe conditions by modulating the noise power, which is scaled linearly with the noise scale factor. 
In this experiment, we confirm that AdaRadar \textit{automatically} reduces the compression ratio under harsher conditions by monitoring the object-detection bounding-box confidence. 
We also compare the results with the vanilla model, which does not employ any compression. 
We find that our compression scheme achieves over 25\texttimes{} compression while maintaining the detection performance in those worse conditions.

\section{Extended quantitative analysis on spectral compression front end}\label{sup:quant_results_fe}

\subsection{Effect of quantization on downstream tasks}

\begin{figure*}[p] 
	\centering
	\includegraphics[width=0.95\textwidth]{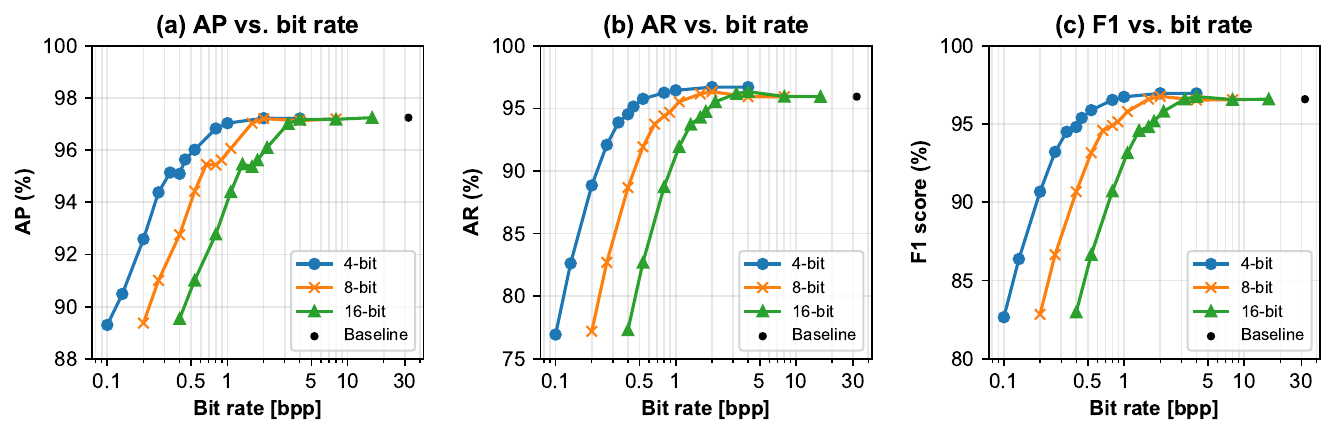}
        \caption{\textbf{Effect of quantization bit width on RADIal.} 
        (a) Average precision, (b) average recall, and (c) F\textsubscript{1} score against the bit rate modulation from pruning.
        We observe that quantization up to 4 bits does not affect the performance compared to that of 8-bit and 16-bit. We use the FFTRadNet on the RADIal dataset \textit{without} any fine-tuning with the block length $M=64$ for all cases.}
	\label{fig:ra_quant_radial}
\end{figure*}

\paragraph{Effect of quantization on RADIal.}
We evaluate how different quantization bit-widths affect both compression–decompression efficiency and network accuracy, and we summarize the resulting design choices.
Figure~\ref{fig:ra_quant_radial} shows the average precision, average recall, and F\textsubscript{1} score roll-off for different bit-rates using the FFTRadNet network on the RADIal \textit{test} dataset \textit{without} any fine-tuning and block length $M=64$.
We observe that 4-bit quantization achieves equal or higher performance compared to bit widths of 8 and 16, and thus we adopt 4-bit quantization.

\begin{figure*}[p] 
	\centering
	\includegraphics[width=0.63\textwidth]{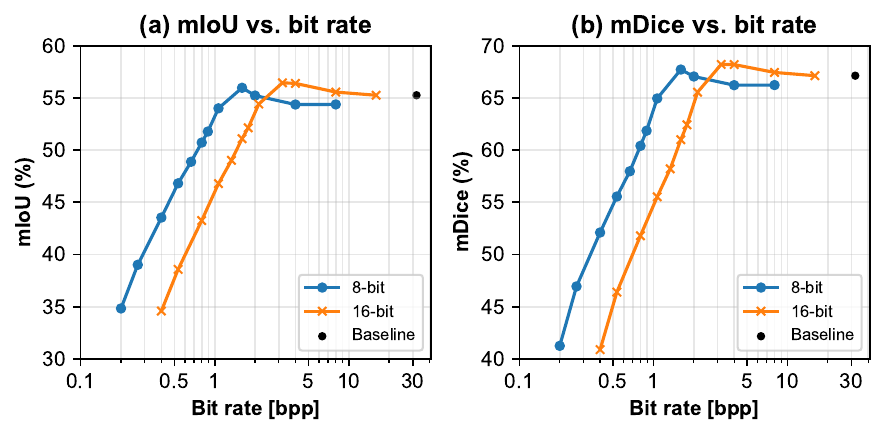}
        \caption{\textbf{Effect of quantization bit width on CARRADA.} 
        (a) mIoU and (b) mDice reported versus the bit rate reduction.
        We observe that quantization up to 8 bits does not affect the performance compared to that of 16-bit. We use the TMVA-Net on the CARRADA dataset \textit{without} any fine-tuning with the block length $M=16$ for all cases.}
	\label{fig:ra_quant_carrada}
\end{figure*}

\paragraph{Effect of quantization on CARRADA.}
Figure~\ref{fig:ra_quant_carrada} depicts how mIoU and mDice vary as the bit-rate decreases, which is the case for the CARRADA dataset with the TMVA-Net.
Performance boosting until 5\texttimes{} pruning is pronounced.
We select 8-bit quantization because it delivers performance similar to that of 16-bit quantization.
We report that 4-bit symmetric quantization underperforms, and we use the block length $M=16$ in the above quantization cases.

\begin{figure*}[p] 
	\centering
	\includegraphics[width=0.95\textwidth]{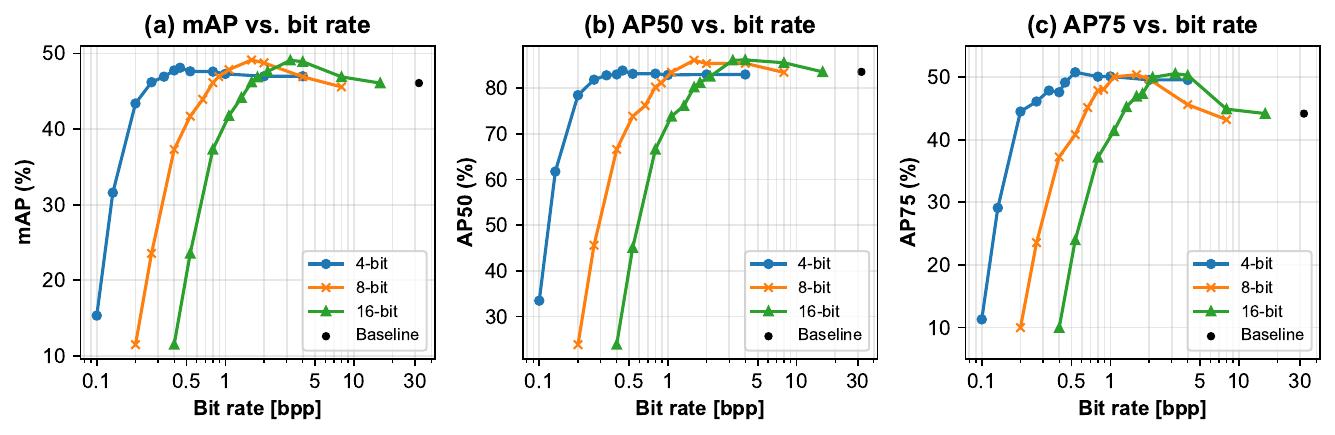}
        \caption{\textbf{Effect of quantization bit width on Radatron.} 
        (a) mAP, (b) AP\textsubscript{50}, and (c) AP\textsubscript{75} vs. the bit rate.
        We observe that quantization up to 8 bits does not affect the performance compared to that of 4-bit and 16-bit. We use the baseline Radatron model on the Radatron dataset \textit{without} any fine-tuning with the block length $M=16$ for all cases.}
	\label{fig:ra_quant_radatron}
\end{figure*}

\paragraph{Effect of quantization on Radatron.}
Figure~\ref{fig:ra_quant_radatron} reveals how detection metrics evolve as the bit-rate is lowered for the Radatron dataset using the baseline Radatron model.
We also observe performance boosting until 7.5\texttimes{} pruning, especially for AP\textsubscript{75}.
We choose 8-bit quantization despite 4-bit quantization giving similar performance due to the consideration for the block length.
We explain this in the following section.

\subsection{Effect of block length on downstream tasks}

\begin{figure*}[p] 
	\centering
	\includegraphics[width=0.95\textwidth]{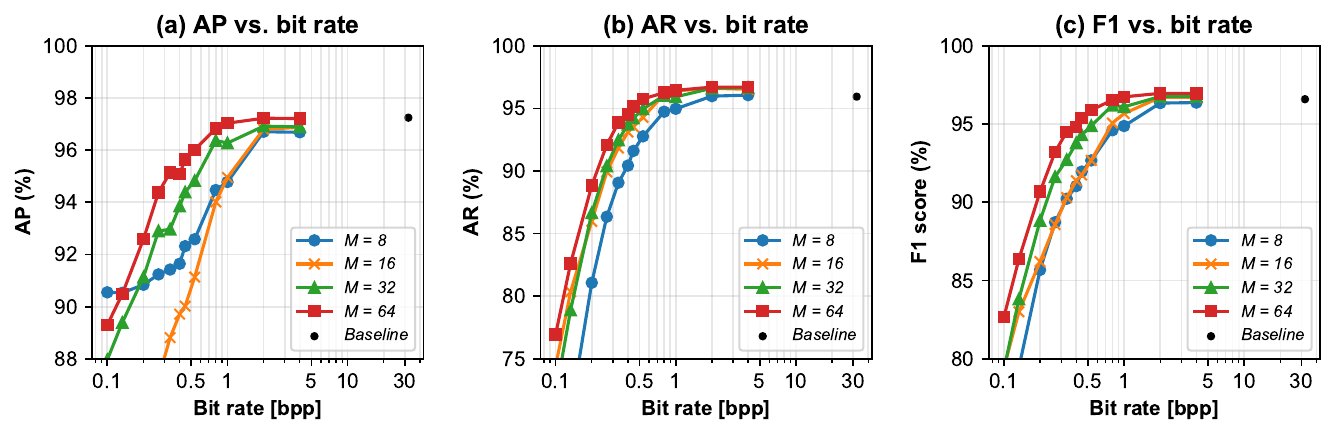}
        \caption{\textbf{Effect of block length on RADIal.} 
        (a) Average precision, (b) average recall, and (c) F\textsubscript{1} score against the block length modulation.
        We observe that compression with a larger block length generally performs better. We use the FFTRadNet on the RADIal dataset \textit{without} any fine-tuning with 4-bit quantization for all cases.}
	\label{fig:ra_BL_radial}
\end{figure*}

\paragraph{Effect of block length on RADIal.}
We examine the impact of varying block sizes on both compression–decompression performance and downstream network accuracy, and draw design choices.
Figure~\ref{fig:ra_BL_radial} presents how average precision, recall, and F\textsubscript{1} score degrade across block sizes $M\in\{8,16,32,64\}$ when evaluated on the RADIal \textit{test} set using FFTRadNet, with 4-bit quantization and no fine-tuning.
Among these, $M=64$ yields the best trade-off between performance and overhead from the scaling factor.

We hypothesize that larger DCT blocks yield better performance because they aggregate more spatial information, leading to smoother and more compressible spectral representations. 
In contrast, smaller blocks may introduce higher variance across patches, making the compression less stable and more lossy.
Also, smaller blocks may experience more fragments on the edges at a high pruning ratio.

\begin{figure*}[p] 
	\centering
	\includegraphics[width=0.63\textwidth]{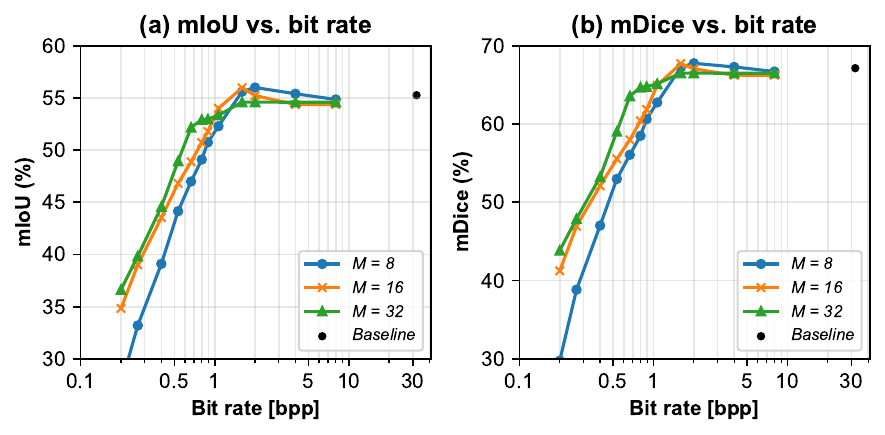}
        \caption{\textbf{Effect of block length on CARRADA.} 
        (a) mIoU and (b) mDice reported versus the change in block length.
        We observe that a larger block is preferred for a higher pruning ratio and a smaller block size for a lower pruning ratio. We use the TMVA-Net on the CARRADA dataset \textit{without} any fine-tuning with 8-bit quantization for all cases.}
	\label{fig:ra_BL_carrada}
\end{figure*}

\paragraph{Effect of block length on CARRADA.}
We evaluate how block length affects segmentation performance on the CARRADA dataset by varying $M\in\{8,16,32\}$ and monitoring mIoU and mDice scores.
Fig~\ref{fig:ra_BL_carrada} shows that $M=32$ offers a favorable trade-off between accuracy and overhead.
Additionally, we report that using $M=64$ without fine-tuning leads to noticeably lower performance.

\begin{figure*}[p] 
	\centering
	\includegraphics[width=0.95\textwidth]{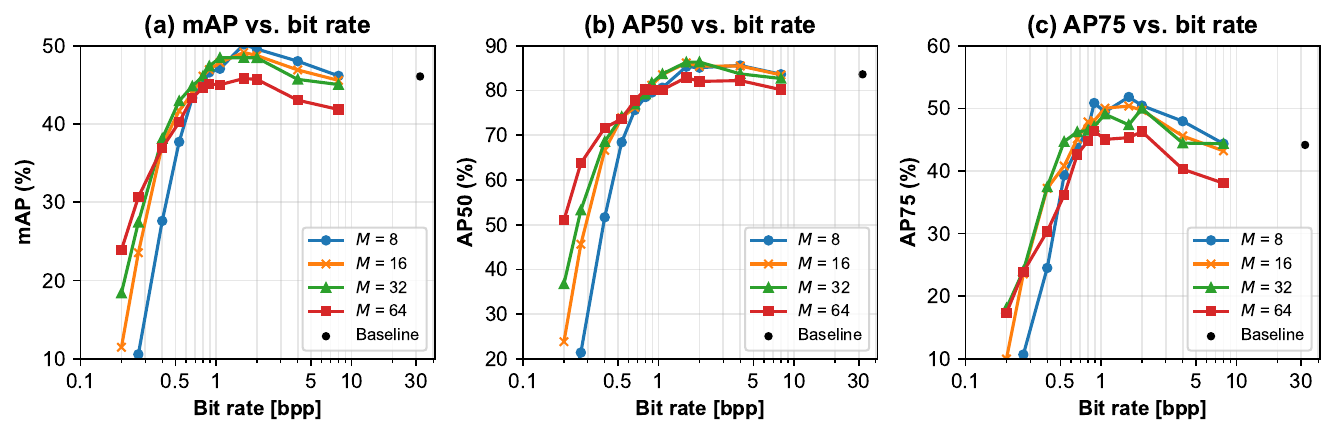}
        \caption{\textbf{Effect of block length on Radatron.} 
        (a) mAP, (b) AP\textsubscript{50}, and (c) AP\textsubscript{75} vs. block size change.
        We observe that a larger patch performs better for higher pruning ratios. We use the baseline Radatron model on the Radatron dataset \textit{without} any fine-tuning with 8-bit quantization for all cases.}
	\label{fig:ra_BL_radatron}
\end{figure*}

\paragraph{Effect of block length on Radatron.}
We assess how varying block sizes influence detection performance on the Radatron dataset by testing $M\in\{8,16,32,64\}$.
Figure~\ref{fig:ra_BL_radatron} indicates that $M=64$ results in relatively poor performance, while $M=32$ provides a more favorable trade-off.

\section{Extended baseline comparisons}\label{sec:ext_base}

\subsection{Comparison with image domain compression methods, Gumbel-sigmoid, and CFAR}

\begin{table*}[h]
\centering
\small  
\caption{\textbf{Benchmark with extended baselines.} We compare radar compression with autoencoders, H.264, Gumbel-sigmoid-based pruning, and CFAR on RADIal.}
\begin{tabular}{cc|cc|ccc|c}
    \toprule
    \textbf{Methods}     & \textbf{Hyperparameters}    & \textbf{BR [bpp]}~$\downarrow$  & \textbf{CR}~$\uparrow$ &  \textbf{Precision} (\%)~$\uparrow$ & \textbf{Recall} (\%)~$\uparrow$ & \textbf{F\textsubscript{1}} (\%)~$\uparrow$ & \textbf{mIoU} (\%)~$\uparrow$ \\
    \midrule
    Baseline    & -                                 & 32        & 1\texttimes{}         & 97.24     & 95.93     & 96.58     & 75.97     \\
    Ours        & $s_\texttt{FxP}=4\text{-bit}$     & 0.32      & 100.5\texttimes{}     & 96.25     & 94.04     & 95.13     & 79.34     \\
    \midrule
    AE          & $\lambda_{ae}=1.0\cdot10^{-3}$    & 1.66      & 19.2\texttimes{}      & 97.20     & 95.67     & 96.43     & 75.45     \\
    AE          & $\lambda_{ae}=3.3\cdot10^{-4}$    & 1.07      & 29.75\texttimes{}      & 97.22     & 94.85     & 96.02     & 74.42     \\
    AE          & $\lambda_{ae}=1.0\cdot10^{-4}$    & 0.59      & 54.1\texttimes{}      & 97.40     & 92.82     & 95.05     & 70.80     \\
    AE          & $\lambda_{ae}=3.3\cdot10^{-5}$    & 0.27      & 117.1\texttimes{}     & 97.49     & 77.05     & 86.08     & 61.04     \\
    \midrule
    H.264       & $\text{crf}=0$                    & 7.8       & 4.08\texttimes{}      & 98.49     & 76.41     & 86.05     & 66.48     \\
    H.264       & $\text{crf}=10$                   & 5.4       & 5.91\texttimes{}      & 98.44     & 76.28     & 85.95     & 66.43     \\
    H.264       & $\text{crf}=20$                   & 3.6       & 8.87\texttimes{}      & 98.33     & 75.76     & 85.58     & 66.26     \\
    \midrule
    Gumbel      & $\lambda_s=0.03$                  & 13.86     & 2.3\texttimes{}       & 96.37     & 96.89     & 96.63     & 75.17     \\
    Gumbel      & $\lambda_s=0.1$                   & 8.98      & 3.6\texttimes{}       & 94.33     & 97.13     & 95.71     & 73.59     \\
    Gumbel      & $\lambda_s=0.3$                   & 6.68      & 4.75\texttimes{}       & 91.24     & 96.23     & 93.67     & 70.40     \\
    Gumbel      & $\lambda_s=1.0$                   & 2.83      & 11.3\texttimes{}      & 84.91     & 90.91     & 87.80     & 57.57     \\
    \midrule
    CFAR        & $thd=10^{0.05}$                   & 7.68      & 4.17\texttimes{}      & 98.14     & 71.54     & 82.76     & 61.10     \\
    CFAR        & $thd=10^{0.2}$                    & 5.56      & 5.76\texttimes{}      & 98.02     & 63.14     & 76.81     & 55.72     \\
    CFAR        & $thd=10^{0.5}$                    & 2.29      & 14.0\texttimes{}      & 98.78     & 42.77     & 59.69     & 47.13     \\
    \bottomrule
\end{tabular}
\label{tab:benchmark}
\end{table*}

\paragraph{Autoencoder.}
Image-domain methods exhibit a performance gap due to the extreme sparsity of radar data, which contrasts with the dense spatial information of optical imagery.
Table \ref{tab:benchmark} compares the compression performance of the autoencoder (AE) architecture by Ballé \etal \cite{balle2017endtoend}, implemented with the CompressAI library \cite{begaint2020compressai}.
The results indicate a substantial $54\times$ reduction in radar data volume.
However, the performance degrades by a larger amount than AdaRadar after compression by two orders of magnitude.

We trained the AE on the RADIal training set for epochs up to 300 with a learning rate of $10^{-4}$ and evaluated its impact by integrating it into the FFTRadNet pipeline.
We treat each radar channel arising from Rx-Tx pairs independently. meaning the channel dimension is flattened with the batch dimension.
The bit rate is measured directly from the bottleneck latent code.
Since networks optimized for specific $\lambda_{ae}$ hyperparameters are trained independently, they are fundamentally \textit{static}. 
Thus, these AEs lack the capacity for dynamic bit-rate modulation during inference.
The AE is comparable in size (3M parameters) to the perception network itself, undermining the objective of lightweight compression.

\paragraph{H.264.}
We evaluate FFTRadNet using raw radar data compressed via the H.264 codec from the ffmpeg library \cite{ffmpeg2024complete} (Table \ref{tab:benchmark}). 
To ensure robust scaling, we clip the radar distribution per channel at the 2nd and 98th percentiles, as standard min-max normalization yields inferior results. 
The compression is implemented by reformatting the radar channel dimension into a temporal sequence of 8-bit grayscale frames.
We observe that H.264-based compression falls short in terms of downstream task performance because raw radar data is highly sparse and exhibits a wide dynamic range.

\paragraph{Soft gradient with Gumbel-sigmoid.}
We employ a Gumbel-sigmoid \cite{jang2016categorical, maddison2016concrete} with $M^2=64^2$ learnable parameters to estimate soft gradients for pruning DCT coefficients. 
In Table \ref{tab:benchmark}, perception performance remains stable up to a compression factor of 2.3\texttimes{}.
We attribute this robustness to the Gumbel-sigmoid acting as a static mask during inference. 
While dynamic pruning via an auxiliary network is possible, it would require transmitting gradients for the entire set of DCT coefficients, thereby posing a communication bottleneck unsuitable for our compression objectives.

We train the Gumbel-sigmoid layer for epochs up to 100 with a learning rate of $10^{-2}$.
During training, the detection backbone remains frozen while we optimize a composite loss $\mathcal{L} = \mathcal{L}_{\text{task}} + \lambda_s \mathcal{L}_{\text{prune}}$ that combines perception task and pruning objectives. 

\paragraph{CFAR.}
Although CFAR \cite{blake1988cfar, farina1986review} facilitates peak detection in radar images, its lossy nature prevents the reconstruction of the raw radar tensor.
We assume that maintaining these peak values provides a viable basis for compression.
We perform compression analogous to the index-value-based method, in which the non-zero elements are determined by the CFAR algorithm rather than by the magnitude alone.
In Table \ref{tab:benchmark}, we use a Cell-Averaging CFAR detector with window and guard widths of $(9,3)$, evaluating peak-based compression across varying SNR thresholds ($thd$).

\subsection{Distribution comparison between image and raw radar tensor}

\begin{figure}[t]
    \centering
    \includegraphics[width=1.0\linewidth]{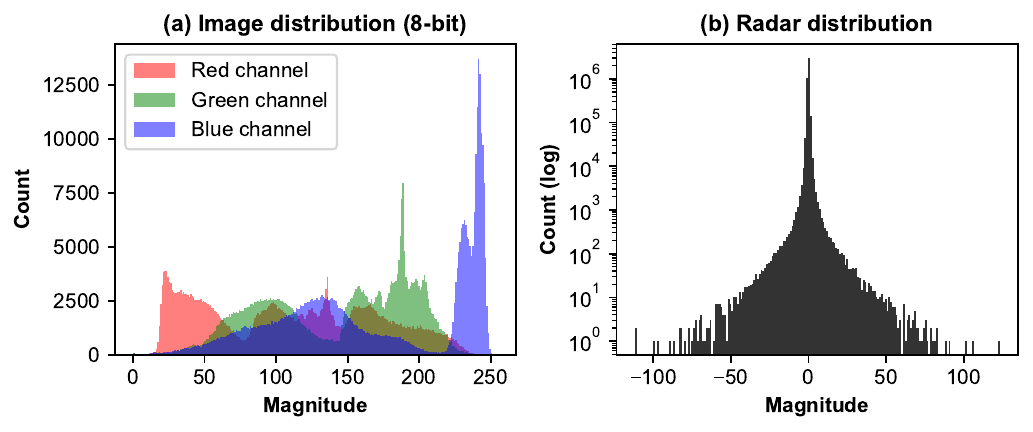}
    \vspace{-1.5em}
    \caption{\textbf{Difference in distribution between image and radar tensors}. This illustrates that while natural (a) images span a broad dynamic range, (b) radar signals exhibit extreme sparsity and a heavy-tailed profile (1st/99th percentiles at $\pm$1.74), necessitating specialized compression beyond standard image-domain codecs.}
    \label{fig:dist_compare}
\end{figure}

\noindent Figure \ref{fig:dist_compare} illustrates the stark contrast between the distributions of natural images and radar tensors. 
While the image data is distributed broadly across the 8-bit dynamic range, the radar data exhibits extreme sparsity. 
Here, the image is of size $3 \times 512 \times 768$ while the radar is $32 \times 512 \times 256$.
Specifically, the radar signal is highly concentrated, with the 1st and 99th percentiles residing at -1.74 and 1.74, respectively. 
This statistical discrepancy highlights why standard image-domain compression codecs are ill-suited for radar data without domain-specific adaptation.

\subsection{Compression after learning-based FFT}

\begin{table*}[h]
    \centering
    \small  
    \caption{\textbf{Compression after learned FFT.} We demonstrate the effectiveness of AdaRadar even with \textit{learned} FFT. We use T-FFTRadNet for detection and segmentation using the RADIal test set. Our compression method achieves 32\texttimes{} data reduction with negligible performance loss. (PR: prune rate; BR: bit rate;  CR: compression rate.)}
    \begin{tabular}{c|cc|cc|ccc|c}
        \toprule
        \textbf{Methods} & \textbf{Bit}~$\downarrow$ & \textbf{PR}~$\uparrow$ & \textbf{BR [bpp]}~$\downarrow$ & \textbf{CR}~$\uparrow$  &  \textbf{Precision} (\%)~$\uparrow$ & \textbf{Recall} (\%)~$\uparrow$ & \textbf{F\textsubscript{1}}(\%)~$\uparrow$ & \textbf{mIoU}(\%)~$\uparrow$ \\
        \midrule
        Baseline \cite{giroux2023t}  & 32 & 1 & 32   & 1\texttimes{} & 90.61 & 93.67 & 92.11 & 80.56 \\
        Ours                        & 4  & 2 & 2    & 16\texttimes{} & 90.94 & 93.19 & 92.05 & 80.34 \\
        Ours                        & 4  & 4 & 1    & 32\texttimes{} & 90.55 & 91.89 & 91.22 & 79.56 \\
        \bottomrule
    \end{tabular}
    \label{tab:learned_fft}
\end{table*}

\noindent We demonstrate the efficacy of AdaRadar on learning-based FFT in Table \ref{tab:learned_fft}.
T-FFTRadNet converts raw ADC inputs into spectral form via FourierNet, which performs \textit{learned} FFT.
Its Swin Transformer backbone uses this spectral representation to perform object detection.
We integrate AdaRadar with 4-bit quantization and patch size $M=8$ after the FourierNet, which performs learnable 2D-FFT from raw ADC outputs.
It compresses the radar tensor by a factor of 32\texttimes{} with negligible loss in accuracy (\textasciitilde{}1\%), \textit{without} retraining.

\section{Qualitative results and visualizations}\label{sup:qual_results}

\subsection{Raw radar data under compression}

\begin{figure*}[h] 
	\centering
	\includegraphics[width=1\textwidth]{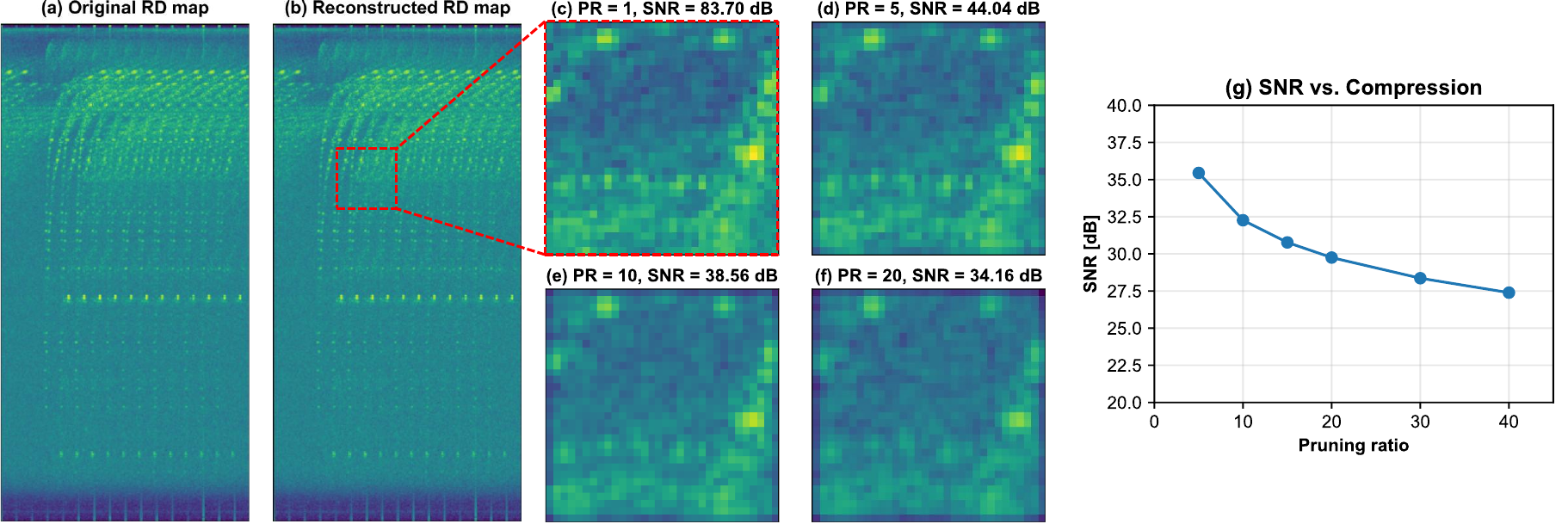}
        \caption{\textbf{Qualitative visualization on range-Doppler map.}
        (a) Original and (b) reconstructed RD map for a single channel with 8-bit quantization and a pruning ratio of 1. 
        (c)-(f) Magnified 64\texttimes{}64 patches at pruning ratios of $\{1,5,10,20\}$, respectively. 
        (g) SNR vs. bit rate trade-off where numerical values are listed in Table \ref{tab:rae}.
        }
	\label{fig:qual_snr}
\end{figure*}

\paragraph{Range-doppler map visualization.}
Figure~\ref{fig:qual_snr} compares the range-Doppler (RD) maps before compression with those reconstructed after the full compression–decompression cycle, and adds magnified patches for each map at several signal-to-noise ratio (SNR) levels.
Figure~\ref{fig:qual_snr}(a) displays the original RD map $\mathbf{x}$, whereas Fig.~\ref{fig:qual_snr}(b) shows its reconstruction $\hat{\mathbf{x}}$ after the compression–decompression process with 8-bit quantization at a pruning ratio of 1.
Each image visualizes the log-power spectrum $\log\|\mathcal{X}_c\| = \log\!\left\lVert \mathbf{x}_{2c} + j\mathbf{x}_{2c+1} \right\rVert$ where the radar’s complex feature tensor $\mathcal{X}\in\mathbb{C}^{C\times H\times W}$ is decomposed into its real and imaginary parts and concatenated along the channel dimension to form the real-valued tensor $\mathbf{x}\in\mathbb{R}^{2C\times H\times W}$.
Figure~\ref{fig:qual_snr}(c) zooms in on a 64\texttimes{}64 patch taken from the reconstructed RD map shown in Fig.~\ref{fig:qual_snr}(b).
Figure~\ref{fig:qual_snr}(d)-(f) respectively correspond to the same patch with different pruning ratios of $\{5,10,20\}$.

\begin{figure*}[p] 
	\centering
	\includegraphics[width=1\textwidth]{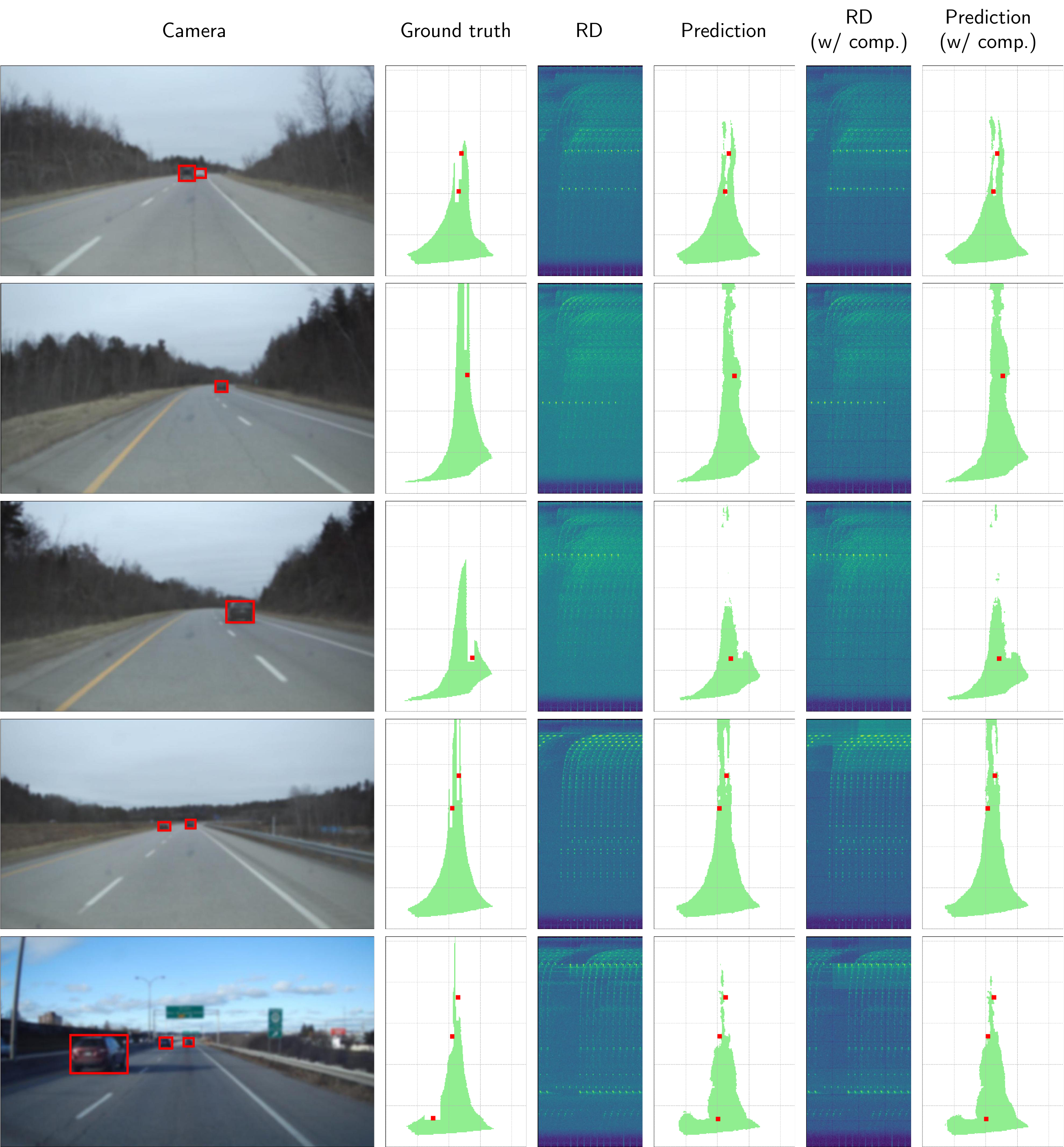}
    \caption{\textbf{Additional qualitative comparisons.} 
    Camera images only provide contextual reference for the scene. RD images map Range to the $y$-axis and Doppler to the $x$-axis. We highlight the detected car in {\color{red} red} and the segmented map in {\color{green} green} in the bird’s-eye-view visualization.}
    \vspace{5em}
	\label{fig:qual_more}
\end{figure*}

Figure~\ref{fig:qual_more} provides additional qualitative results to showcase the effect of compression on the RD map and the corresponding radar perception outputs.
Results on arbitrary samples consistently indicate that object detection or segmentation outputs perform comparably to the non-compressed baseline.

\begin{table}[h]
    \centering
    \small
    \caption{\textbf{SNR \& RAE against compression.}
    Mean and maximum relative absolute errors (RAE) at different compression ratios.}
    \begin{tabular}{c c c c}
    \toprule
    \textbf{Comp. Ratio} $\uparrow$ & \textbf{SNR [dB]} $\uparrow$ & \textbf{RAE}\textsubscript{mean} $\downarrow$ & \textbf{RAE}\textsubscript{max} $\downarrow$ \\
    \midrule
    1       &  144.17   &   0.000   &   0.000 \\
    5       &   35.45   &   0.010   &   0.329 \\
    10      &   32.26   &   0.017   &   0.378 \\
    15      &   30.77   &   0.022   &   0.389 \\
    20      &   29.75   &   0.025   &   0.390 \\
    30      &   28.36   &   0.031   &   0.383 \\
    \bottomrule
    \end{tabular}
    \label{tab:rae}
\end{table}

\paragraph{SNR against compression.}
Figure~\ref{fig:qual_snr}(g) plots the average SNR against bit-rate for 100 RD maps sampled from the RADIal \textit{test} set.
The SNR falls steeply as the pruning ratio rises from 1\texttimes{} to 5\texttimes{}. 
The curve flattens beyond 5\texttimes{}, indicating only marginal degradation.

\paragraph{RAE against compression.} 
We computed the mean and maximum relative absolute errors (RAE) between the original and the reconstructed power spectrums (Table \ref{tab:rae}). Here, $\text{RAE} = \frac{|x - \hat{x}|}{|x|}$ where $x$ and $\hat{x}$ respectively are the original and reconstructed spectrums, and both mean and max RAE are averaged across 100 radar tensor samples from the RADIal test set. While the maximum RAE saturates at approximately 0.38 beyond the compression ratio of 10, the mean RAE grows sub-linearly with a greater compression ratio. These results demonstrate the robustness and effectiveness of our compression method in preserving spectral fidelity even at high compression rates.

\subsection{Remarks on frequency components}\label{sec:dct_remark}

\begin{figure}[h]
    \centering
    \begin{subfigure}[t]{0.48\linewidth}
        \centering
        \includegraphics[width=\linewidth]{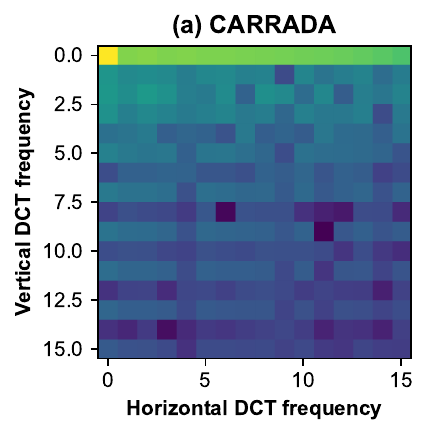}
    \end{subfigure}
    \hfill
    \begin{subfigure}[t]{0.48\linewidth}
        \centering
        \includegraphics[width=\linewidth]{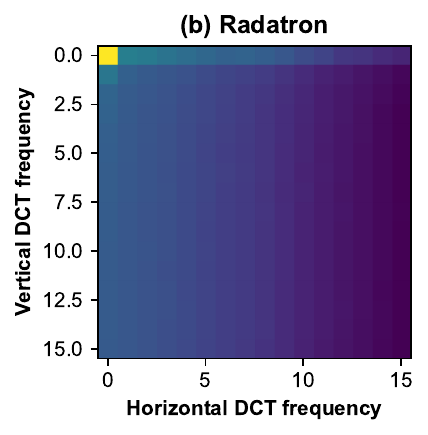}
    \end{subfigure}
    \caption{\textbf{DCT coefficient magnitudes.}
    (a) CARRADA and (b) Radatron datasets. Unlike the RADIal dataset, the spectral bins are more focused on low-frequency bins. Our adaptive pruning is agnostic to such spectral structures, since it ranks coefficients by energy.
    }
    \label{fig:noise_levels}
\end{figure}

\noindent Our method \textit{adaptively} selects coefficients within each patch based on local energy, making it broadly applicable across different sensor types and downstream tasks.
The frequency components are determined primarily by the radar sensor characteristics and the perceived scene.
We provide further remarks on the distribution of frequency coefficients in the 2D map.

\paragraph{RADIal.} In Figure~\ref{fig:DCT_coef}(a), when we experiment with radar tensors from the RADIal dataset, the magnitude of the DCT coefficients is concentrated around the high-frequency components, with peak energy at $(x,y)=(63,63)$ and decreasing radially. Taking a horizontal slice at $y=63$ or a vertical slice at $x=63$, we observe a linearly increasing ramp in frequency towards the end. 

\paragraph{CARRADA.} We observe that the energy is centered mostly at the low-frequency components, exhibiting high-intensity values at the left and top edges, i.e., $x=0$ and $y=0$. This suggests that the radar sensors capture smooth horizontal and vertical patterns in the range Doppler map.

\paragraph{Radatron.} The pattern is also similar for the Radatron dataset, wherein we discover the energy mainly being concentrated in the low-frequency bins, creating high-intensity bar shapes at the left and top edges.


\medskip




\end{document}